\documentclass[acmsmall,screen]{acmart}
\AtBeginDocument{%
  }

\usepackage{amsmath,amsfonts}
\usepackage{algorithmic}
\usepackage{algorithm}
\usepackage{array}

\usepackage{textcomp}
\usepackage{stfloats}
\usepackage{url}
\usepackage{verbatim}
\usepackage{graphicx}
\usepackage{subcaption}
\usepackage[most]{tcolorbox}

\usepackage{booktabs}
\usepackage{tabularx}
\usepackage{supertabular}

\begin{document}

\title{AgentCAT: Simulating Computerized Adaptive Testing via Multi-Agent Large Language Models}


\author{Weiyuan Zhou}
\affiliation{%
  \institution{State Key Laboratory of Opto-Electronic Information Acquisition and Protection Technology, Institute of Physical Science and Information Technology, Anhui University}
  \city{Hefei}
  \state{Anhui}
  \country{China} 
}
\email{zwy8147@gmail.com}

\author{Haiping Ma}
\authornote{Corresponding Author.}

\affiliation{%
  \institution{State Key Laboratory of Opto-Electronic Information Acquisition and Protection Technology, Institute of Physical Science and Information Technology, Anhui University}
  \city{Hefei}
  \state{Anhui}
  \country{China}
}
\email{hpma@ahu.edu.cn}

\author{Xiaoshan Yu}
\affiliation{%
  \institution{School of Artificial Intelligence, Anhui University}
  \city{Hefei}
  \state{Anhui}
  \country{China}
}
\email{yxsleo@gmail.com}

\author{Changqian Wang}
\affiliation{%
  \institution{School of Computer Science and Technology, Dalian University of Technology}
  \city{Dalian}
  \state{Liaoning}
  \country{China}
}
\email{changqian.wang.dl@gmail.com}

\author{Shangshang Yang}
\affiliation{%
  \institution{School of Computer Science and Technology, Anhui University}
  \city{Hefei}
  \state{Anhui}
  \country{China}
}
\email{yangshang0308@gmail.com}


\author{Xingyi Zhang}

\affiliation{%
  \institution{School of Computer Science and Technology, Anhui University}
  \city{Hefei}
  \state{Anhui}
  \country{China}
}
\email{xyzhanghust@gmail.com}
\renewcommand{\shortauthors}{Trovato et al.}


\begin{abstract}
Computerized Adaptive Testing (CAT), as a key technology for realizing personalized education, aims to accurately assess an examinee's proficiency by retrieving exercises from a candidate pool that dynamically match their current ability estimates. However, existing CAT research paradigms have long been constrained by the dual limitations of static offline data and isolated component optimization. Severely restricted by the issue of partial labels in offline interaction logs, researchers are forced to degrade the inherently dynamic assessment process into a static sequence prediction problem. Meanwhile, current research predominantly focuses on isolated perspectives—such as either the selection strategy or the diagnostic model—neglecting the holistic investigation of the overall CAT interaction process. To break through these bottlenecks, this paper proposes AgentCAT, a Large Language Model-based multi-agent simulation system, aiming to construct a high-fidelity benchmarking simulation environment for dynamic testing. This framework comprises three core modules: (1) The examinee agent, equipped with memory retrieval and Chain-of-Thought reasoning capabilities, acts as a cognitive simulator to replicate authentic response behaviors based on cognitive profiles; (2) The selection agent acts as an instructional planner, employing a coarse-to-fine bucketing strategy and a knowledge graph exploration mechanism to strike a balance between local difficulty matching and global knowledge coverage; (3) The supervisor introduces a dual-auditing and robust update mechanism to ensure the convergence and statistical validity of ability estimation. To comprehensively validate the effectiveness of the framework, we conducted systematic evaluations on two real-world datasets across three dimensions: macro-level ability convergence, micro-level instructional interaction logic, and the capability to overcome data sparsity. Experimental results demonstrate that AgentCAT not only achieves highly effective ability estimation, but its selection strategy also successfully strikes a balance between difficulty adaptation and instructional coherence that perfectly aligns with human pedagogical intuition.
\end{abstract}




\maketitle

\section{Introduction}

\begin{figure}[htbp]
\centering
\setlength{\abovecaptionskip}{0.cm}
\includegraphics[scale=0.18]{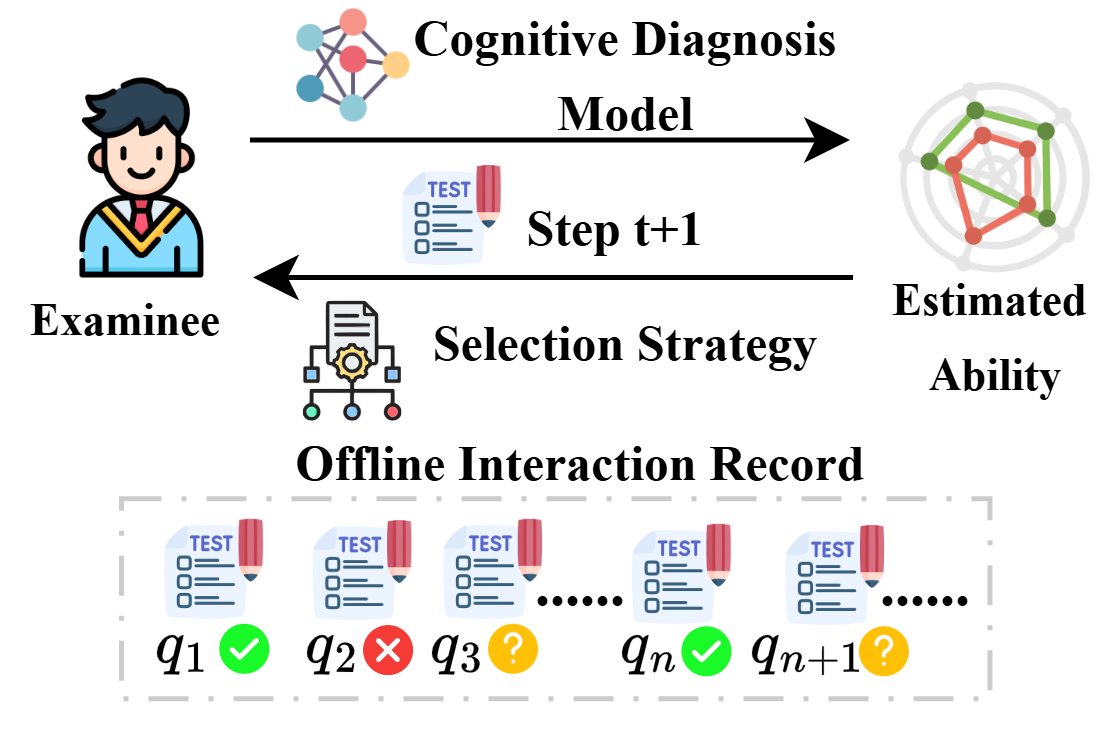}
\caption{A schematic illustration of a typical CAT process relying on static offline data with missing ground-truth labels.}
\label{CAT_fig}
\end{figure}

Computerized Adaptive Testing (CAT) is one of the core technologies for realizing personalized proficiency assessment in intelligent education. Its central philosophy is to model the examinee's proficiency assessment process as a dynamic interactive closed-loop, aiming to achieve maximum precision in evaluating the examinee's ability within a finite number of interactions by sensing their behavior in real-time~\cite{bi2024model}. Diverging from the traditional ``one-size-fits-all'' fixed-form testing, CAT, as a sequential pedagogical decision-making system, can dynamically adjust the difficulty of exercises based on the examinee's real-time feedback, thereby truly realizing a tailored paradigm of personalized assessment~\cite{zhuang2026survey}. To achieve these objectives, CAT systems typically rely on the deep coupling of two core components: the Cognitive Diagnosis Model~\cite{guodiagnosis}~(CDM)~and the Selection Strategy. As illustrated in Figure~\ref{CAT_fig}, the former is responsible for inferring the examinee's current knowledge proficiency based on their historical response records; the latter serves as a pedagogical decision-making engine, utilizing specific strategies based on these assessment results to retrieve the optimal exercise for step $t+1$ from the exercise pool. This ``diagnosis-selection-response'' closed-loop process iterates continuously until pre-set termination criteria are met, ultimately yielding a comprehensive cognitive profile of the examinee.

Despite significant progress in CAT research, existing CAT research paradigms are universally constrained by the dual shackles of static offline data and isolated component optimization. Specifically, current CAT research typically relies entirely on offline datasets published by real-world online education systems~\cite{zhuang2026survey}. However, in authentic educational scenarios, a single examinee often interacts with only a minute fraction of the exercises within a massive pool. This phenomenon forces researchers to confront the severe challenge of partial labels in offline interaction logs, where offline interaction records exhibit extreme sparsity and incomplete knowledge coverage~\cite{yao2023exploiting}. Consequently, researchers are compelled to conduct static back-testing of the system solely along predetermined and fixed historical interaction trajectories. This evaluation approach, which degrades into static sequence prediction, causes the CAT system to deviate from its interactive essence as a dynamic pedagogical system~\cite{zhuang2022fully}: a minor modification in the selection strategy would directly lead the system into an entirely different interaction path and alter the convergence trajectory of the CDM in real-time. Moreover, this implies that the diagnostic performance of CAT across the vast majority of knowledge concepts remains fundamentally unknown, thereby failing to adequately profile the true proficiency of the examinee~\cite{ma2024enhancing}. Concurrently, existing CAT research predominantly adopts a singular perspective: fixing either the CDM or the selection strategy while optimizing the other as an isolated module (e.g., exploring how to accelerate the early-stage convergence of the CDM~\cite{liu2025fast} or how to enhance the precision of the selection algorithm~\cite{wang2025explicit}). This fragmented research paradigm neglects the complex dynamic coupling effects between exercise selection and cognitive diagnosis, making it difficult for experimental results to authentically reflect the system's overall performance, instructional coherence, and robustness in open educational scenarios. Confronted with these dilemmas, constructing a high-quality pedagogical simulation environment capable of supporting free interaction and real-time feedback has become an essential path to break through existing research bottlenecks and restore the dynamic essence of CAT.

Against this backdrop, the remarkable human-like reasoning and role-playing capabilities demonstrated by Large Language Models (LLMs) in recent years have provided a revolutionary technical foundation for constructing such high-fidelity educational simulation environments~\cite{11410595,naveed2025comprehensive,lang10669809}. Existing studies indicate that LLM-based examinee agents are capable of generating cognitively plausible response behaviors under given profiles~\cite{gao2025agent4edu}. This renders the use of agents to simulate examinees for full-pipeline interaction a viable pathway to break through the constraints of static data. However, most current works predominantly treat the examinee agent as an independent data generator, having not yet systematically integrated it into the decision-making closed-loop of CAT. To this end, this paper proposes \textbf{AgentCAT}, a novel CAT research framework. Rather than being confined to proposing a specific selection strategy, this paper leverages the superior simulation and reasoning capabilities of agents to reconstruct the CAT process into a multi-agent collaborative simulation system, aiming to simulate the entire interactive closed-loop of adaptive testing. Realizing this framework requires addressing three core challenges: (1) Cognitive Simulation Fidelity: How to leverage the rich corpus knowledge and reasoning capabilities of LLMs to authentically simulate an examinee's response behaviors?
(2) Pedagogical Strategy Alignment: How to harness the pedagogical reasoning and planning capabilities of LLMs to orchestrate a multi-objective selection strategy—balancing difficulty matching, knowledge coverage, and remedial instruction—within a large-scale exercise pool?
(3) Dynamic Ability Calibration: How to mitigate the generative uncertainty (or hallucinations) of LLMs to ensure the accuracy and robustness of ability estimation during dynamic multi-agent interactions?

Therefore, AgentCAT models the CAT process as a multi-agent interactive process. Specifically, AgentCAT abandons the traditional mechanical abilty asssessment paradigm and constructs a dynamic interactive system comprising multiple decision-making roles, which includes three key components:  (1) The examinee agent: Constructed based on examinee profiles from real-world datasets, this agent internally maintains a specific latent knowledge state. It leverages the reasoning capabilities of LLMs to simulate the authentic response behaviors of examinees when encountering exercises of varying difficulties, thereby providing a dynamic simulated interactive environment for the system. (2) The selection agent: Acting as a pedagogical decision-maker with a global perspective, it integrates educational psychology principles such as prerequisite dependency and the Zone of Proximal Development (ZPD)~\cite{cai2025exploring}. This ensures that the testing sequence not only achieves extremely high diagnostic efficiency but also maintains robust knowledge coherence and instructional logic. (3) The supervisor: This module incorporates a hybrid response and asynchronous soft-overwriting mechanism grounded in real-world data anchors. It monitors the interaction trajectory between the two agents in real-time, utilizing ground-truth labels as correction signals. This ensures that while granting LLMs ample freedom for exploration, the system's final ability estimation strictly converges to the statistical laws of real-world educational assessment.

Unlike traditional CAT methodologies, the core contribution of AgentCAT lies not in proposing a specific diagnostic model or selection formula, but rather in establishing a controllable, interpretable, and scalable CAT simulation system. This system is capable of reproducing the complete dynamic interactive assessment process with high fidelity without the participation of real online examinees, serving as a novel data-anchored experimental platform for the analysis, benchmarking, and validation of future adaptive testing methodologies.

\section{Related Work}

\subsection{Computerized Adaptive Testing}
Despite substantial advancements in CAT in recent years, existing research has largely failed to transcend the constraints of traditional paradigms, predominantly clustering into two dimensions: functional optimization of isolated components and adaptive remediation under static data. From the data perspective, to address the cold-start and dynamic adaptation challenges posed by static offline data, researchers have attempted remediation by introducing external priors\cite{ma2025diffusion} or Reinforcement Learning (RL) mechanisms\cite{wang2023gmocat,yu2025light}. Due to the absence of counterfactual interaction records in offline datasets, traditional CAT often struggles to converge rapidly during the initial testing phase (cold-start stage). Addressing this data sparsity issue, DCSR\cite{ma2025diffusion} proposes a cross-domain transfer framework based on diffusion models, establishing a bridge for cognitive state transfer across domains to provide high-quality prior distributions for initial ability estimation. FACD\cite{liu2025fast} focuses on dynamic adaptability in the early stages, designing a fast adaptive cognitive diagnosis framework that utilizes personalized sequence modeling to construct precise examinee profiles within minimal interaction turns. Furthermore, NCAT \cite{zhuang2022fully} and GMOCAT \cite{wang2023gmocat} incorporate RL mechanisms, modeling CAT as a bi-level optimization problem\cite{ijcai2021p332}. By utilizing Q-learning or Actor-Critic algorithms to continuously learn from offline interaction data, these methods capture non-linear relationships, thereby enhancing assessment precision. However, even with these end-to-end optimization methods, their training and evaluation remain constrained by static historical response logs, failing to truly simulate the dynamic evolutionary behavior of new strategies under counterfactual paths.~From the component optimization perspective, extensive research focuses on enhancing the efficiency or precision of the selection strategy as a standalone module \cite{wang2025explicit,ma2025reconciling}, often overlooking its dynamic coupling with the diagnostic model. In the early stages of research, statistical methods based on Fisher Information \cite{lord2012applications} and Kullback-Leibler (KL) Divergence \cite{chang1996global} were widely adopted due to their theoretical rigor; however, these approaches exhibit a high dependency on specific CDM. To transcend this limitation, researchers have begun to explore model-agnostic universal strategies. Notably, the MAAT \cite{bi2020quality} method has garnered significant attention by introducing an Active Learning perspective. It dynamically maximizes information gain by calculating the expected model change of candidate questions, while simultaneously balancing knowledge coverage during the selection process.

\subsection{LLM-based Educational Agents}
With the breakthroughs of  LLMs in open-domain knowledge acquisition and complex logical reasoning, the research paradigm of intelligent agents is shifting from the traditional mode of adapting to closed environments via policy learning to an autonomous agent paradigm with the LLMs serving as the core controller\cite{vill10648793,yang11192789}.

Characterized by a high emphasis on individual differences and continuous interaction, the field of intelligent education is regarded as an ideal scenario for deploying agents\cite{xi2025investigating,wang2025llm}. By integrating tool-use, memory mechanisms, and planning capabilities, these educational agents play an increasingly pivotal role within the educational ecosystem\cite{yu2025simulated,HaoCLYLZ26,ZhengSG26,XiZW26}. Based on the target audience, existing research on educational agents can be primarily categorized into two classes\cite{chu-etal-2025-llm}: (1) Teacher-facing Instructional Assistant Agents: These agents aim to alleviate teacher burden and enhance instructional design capabilities through automated and intelligent means. They can assist teachers not only in routine tasks such as lesson plan generation\cite{hu2025exploring} and assignment grading but also in optimizing teaching strategies by simulating complex classroom environments\cite{zhang2025simulating}. Such applications not only provide a low-cost training ground for teachers but also demonstrate the potential of LLMs agents in simulating group teaching behaviors. (2) Examinee-facing Personalized Learning Agents: These agents are dedicated to providing round-the-clock adaptive tutoring and domain-specific skill training for examinees. They are capable of dynamically adjusting feedback strategies based on the examinee's historical performance or providing in-depth support in specific disciplines\cite{rogers2025playing,NguyenPGTS25}. This strategy based on role simulation not only improves character consistency in stories but also effectively helps examinees master complex narrative skills, highlighting the unique advantages of agents in cultivating specific skills\cite{wei2025igniting,fu11204697}. The aforementioned studies demonstrate that agents now possess advanced capabilities to simulate teachers and examinees respectively. Introducing the paradigm of multi-agent collaboration\cite{qian2025scaling} to integrate these two types of agents not only holds significant engineering value but also lays a foundation for constructing a CAT simulation system that combines theoretical rigor with practical feasibility.

However, constructing a multi-agent-based CAT simulation system is by no means a simple stacking of dialogues between a Selection Agent and an Examinee Agent\cite{zhu2025multiagentbench}. Realistic CAT simulations confront far more stringent mathematical constraints than general instructional dialogues; the core challenge lies in the precise estimation of ability during interaction. During the simulation process, the two agents generate a vast amount of natural language interactions and behavioral decisions\cite{arana2025foundations}. How to filter out LLMs hallucinations and stochastic noise from these unstructured interaction data, and leverage psychometric models to accurately quantify and update the examinee's ability state in real-time, constitutes the critical bottleneck in realizing the closed loop of CAT simulation.

\section{Preliminaries}
\subsection{Computerized Adaptive Testing}
In intelligent education scenarios, CAT efficiently assesses examinee proficiency within a finite test length by dynamically adjusting question selection based on response behaviors through continuous dynamic interactions. Formally, let the set of examinees be $\mathcal{S} = \{s_1, s_2, s_3, \dots, s_{|\mathcal{S}|}\}$ and the question bank be $\mathcal{Q} = \{q_1, q_2, q_3, \dots, q_{|\mathcal{Q}|}\}$. An interaction between an examinee and a question can be represented as a triplet $(s, q, y)$, where $y \in \{0, 1\}$ denotes the response outcome of examinee $s$ to question $q$ (1 indicates a correct answer, 0 otherwise). In the CAT process, it is assumed that the examinee's true ability remains invariant during the test. A standard CAT process consist off two core components: the CDM and Selection Strategy. At the stet $t$ , the selection strategy chooses the next question $q^{(t)}$ from  the candidate question bank on the current ability estimate $\hat{\theta}^{(t-1)}$ and the set of available questions. Subsequently, the ability estimate is updated based on the examinee's response feedback. This interactive processes can formally defined as follow:
\begin{align}
    q^{(t)} = \mathcal{P}(\hat{\theta}^{(t-1)}, \mathcal{Q}_s), \quad \hat{\theta}^{(t)} = \mathcal{D}(\hat{\theta}^{(t-1)},q^{(t)},y^{(t)}),
\end{align}
where $\mathcal{Q}_s \subset \mathcal{Q}$ denotes the set of questions not yet answered by examinee $s$,   $\hat{\theta}^{(t)}$ denote the ability estimates at  step $t$, $\mathcal{P}(\cdot)$ is the selection strategy, and $\mathcal{D}(\cdot)$ is the CDM. The two modules operate alternately during the testing process until the maximum number of test step $T$ is reached, at which point the CAT assessment process terminates.

\begin{figure*}[htbp] 
    \centering
    
    \includegraphics[width=1\textwidth]{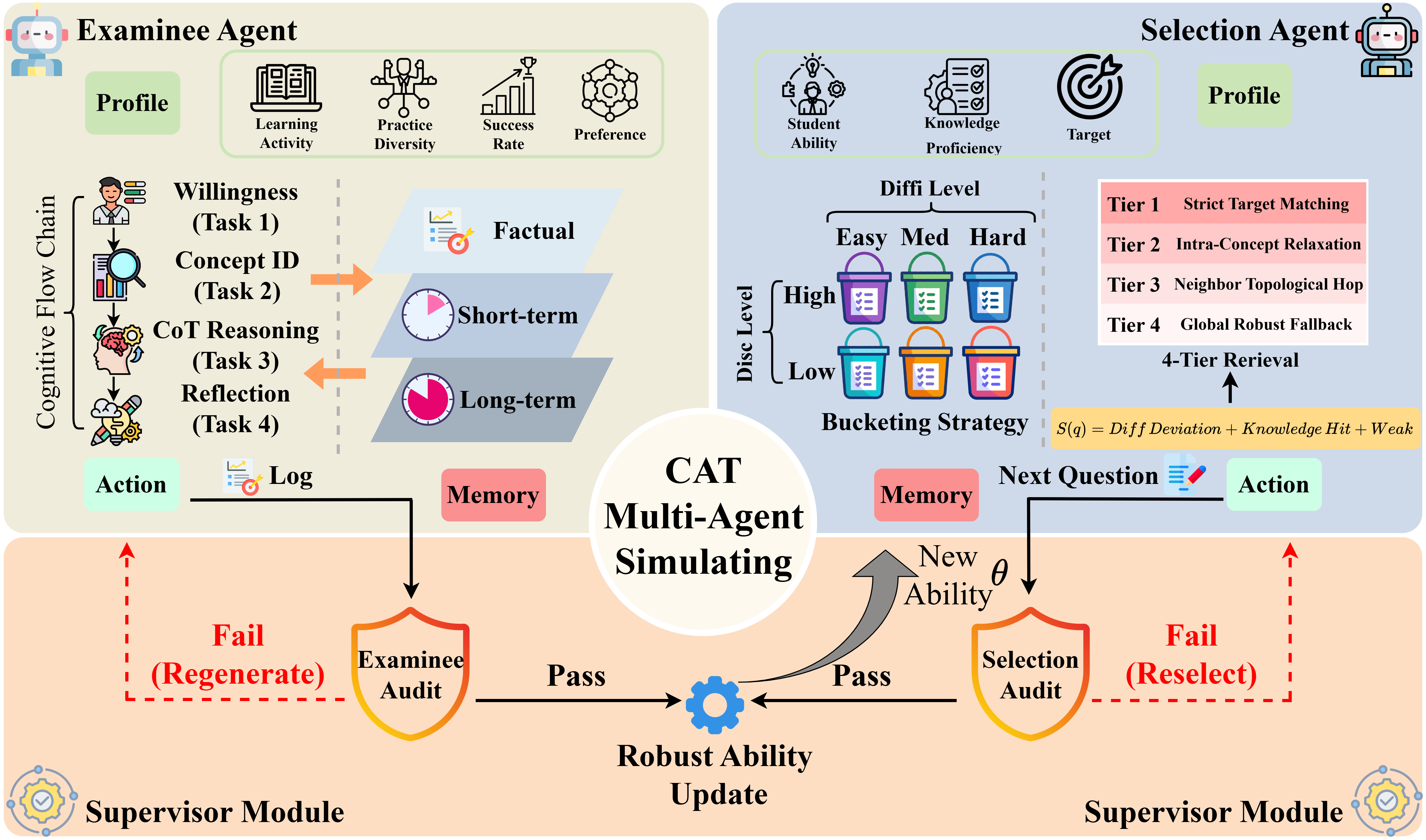}
    
    \caption{Illustration of AgentCAT framework. The AgentCAT framework consists of three components: the Examinee Agent, the Selection Agent, and the Supervisor Module. The Examinee Agent is responsible for simulating examinee response behaviors, the Selection Agent selects questions based on response outcomes and ability estimates, and the Supervisor Module monitors the outputs of both agents and the ability estimation process.}
    \label{framework}
\end{figure*}
\vspace{-1em}

\begin{algorithm}[!t]
\caption{AgentCAT Interaction Process}
\label{alg:agentcat}
\begin{algorithmic}[1]
\REQUIRE Question Bank $\mathcal{Q}$, Knowledge Concept Graph $\mathcal{G}$, Examinee Profile $\mathcal{P}$, Max Step $T$
\ENSURE Final Estimated Ability $\hat{\theta}^{(T)}$

\STATE \textbf{Initialize:} Estimated ability $\hat{\theta}^{(0)} \leftarrow 0$, History $\mathcal{H}_0 \leftarrow \emptyset$
\STATE \textbf{Construct:} Initialize Examinee Agent $\mathcal{A}_{exa}(\mathcal{P})$ , Selection Agent $\mathcal{A}_{sel}(\mathcal{H}_0 )$ and Supervisor Module $\mathcal{V}(\hat{\theta}^{(0)})$

\FOR{$t = 1$ to $T$}
    \STATE \textit{// Phase 1: Coarse-to-Fine Selection (See Section 4.3)}
    \IF{$t = 1$}
        \STATE $q_t \leftarrow \textsc{RandomSelect}(\mathcal{Q})$ \quad \textit{// Cold Start Strategy}
    \ELSE
        \STATE $q_t \leftarrow \mathcal{A}_{sel}.\textsc{Select}(\hat{\theta}_{t-1}, \mathcal{H}_{t-1}, \mathcal{G}, \mathcal{Q})$
    \ENDIF
    
    \STATE \textit{// Phase 2: Cognitive Simulation (See Section 4.2)}
    \STATE $y_{agent}, conf_t \leftarrow \mathcal{A}_{exa}.\textsc{SimulateStep}(q_t)$
    
    \STATE \textit{// Phase 3: Supervision and Update (See Section 4.4)}
    \STATE $y_{\text{truth}} \leftarrow \mathcal{V}.\textsc{LookupGroundTruth}(q_t)$
    \STATE $y_t \leftarrow \mathcal{V}.\textsc{TruthFusion}(y_{agent}, y_{\text{truth}}, conf_t)$
    \STATE $\hat{\theta}^{(t)} \leftarrow \mathcal{V}.\textsc{UpdateAbility}(\hat{\theta}^{(t-1)}, q_t, y_t)$
    
    \STATE \textbf{Update History:} $\mathcal{H}_t \leftarrow \mathcal{H}_{t-1} \cup \{(q_t, y_t)\}$
\ENDFOR

\RETURN $\hat{\theta}^{(T)}$
\end{algorithmic}
\end{algorithm}

\section{Methodology}

\textbf{Overview.} In this section, we formally present the AgentCAT framework. We first outline the overall workflow of multi-agent collaboration within AgentCAT. Subsequently, we detail three core components: the Examinee Agent for simulating examinee response behaviors, the Selection Agent for knowledge-concept-aware question retrieval, and the Supervisor Module for robust ability estimation. It is worth noting that our primary focus lies on the Selection Agent and the Supervisor Module, which represent extended research built upon the foundation of existing the Examinee Agent work.

\subsection{The AgentCAT Framework}
Distinct from traditional computerized adaptive testing methods that rely on statistical metrics or are data-driven but confined to static offline data, AgentCAT models the adaptive testing process as a multi-turn closed-loop interaction among agents. The framework is shown in the Figure\ref{framework}. The framework comprises three main entities: the Selection Agent ($\mathcal{A}_{sel}$) acting as the examiner, the Examinee Agent ($\mathcal{A}_{exa}$) acting as the examinee, and the Supervisor Module ($\mathcal{V}$) acting as the auditor. Formally, let $\mathcal{S}$ denotes the target examinee set, $\mathcal{Q} $ denotes the question bank, and $\mathcal{K} = \{k_1, k_2, k_3, \dots, k_{|\mathcal{K}|}\}$ denotes the set of knowledge concepts, where each question $q_j$ is characterized by a tuple $(c_j, k_j, a_j, b_j)$, representing the question text content, knowledge concept, discrimination parameter, and difficulty parameter, respectively. The relationships between knowledge concepts are defined by a Knowledge Concept Graph $\mathcal{G}$. The entire CAT process proceeds over discrete time steps $t=1, 2, \dots, T$. At each step $t$, the system maintains an estimated latent ability value $\hat{\theta}^{(t)}$ and a historical interaction set $\mathcal{H} = \{(q_1, y_1), (q_2, y_2), \dots, (q_t, y_t)\}$, where $y \in \{0, 1\}$ is the label indicating whether the response is correct (recorded as 1 for correct, 0 otherwise). The overall workflow is illustrated in Algorithm\ref{alg:agentcat}.

\subsection{Examinee Agent}
To address the inherent data sparsity challenges in static educational datasets and enable counterfactual evaluation on unobserved questions, we employ a generative Examinee Agent as a digital twin of real examinees. Built upon the Agent4Edu\cite{gao2025agent4edu} architecture, our Examinee Agent transcends traditional probabilistic generative models, constructing a cognitive simulator that mimics the full lifecycle of learning and problem-solving. This agent is synergistically driven by three integrated modules: Profiling, Memory, and Action. The Profiling Module is responsible for initializing the agent's inherent characteristics, including learning activity, diversity preferences, success rate and initial latent ability. These features are extracted from historical interaction data via the IRT model, ensuring the behavioral consistency of the simulated persona. The Memory Module implements a multi-level storage mechanism inspired by human cognitive theories: Sensory Memory buffers current question inputs; Short-term Memory maintains recent interaction contexts; and Long-term Memory stores consolidated knowledge proficiency and key learning facts. Crucially, the Action Module discards the black-box mode of directly generating binary labels, instead explicitly simulating the mental flow of problem-solving through a Cognitive Task Chain. For a given question, the agent sequentially executes: (1)~Willingness Assessment: Deciding whether to attempt the question based on perceived difficulty. (2)~Concept Identification: Parsing the underlying knowledge concepts behind the question. (3)~Reasoning Generation: Outputting a solution path in textual form. (4)~Self-Reflection: Predicting response correctness and providing a confidence signal. This mechanism endows AgentCAT with the unique capability to explore examinee performance within the counterfactual space where ground-truth labels are absent.

\subsection{Selection Agent}
The Selection Agent serves as the decision-making core of AgentCAT, aiming to recommend the question with the highest diagnostic value at each step. Unlike traditional methods that greedily maximize statistical information gain, AgentCAT confronts two unique engineering challenges: the real-time retrieval latency in large-scale question banks, and the context window constraints of LLMs, which prevent effective perception of information across the entire bank. To address these issues, we discard the brute-force search pattern over the entire question bank $\mathcal{Q}$ and propose a bucket-based ‘‘ Coarse-to-Fine ’’ selection mechanism\cite{jiang2023active}. This mechanism aims to balance the depth of semantic reasoning with computational real-time performance within a massive search space through structured index preprocessing.
\subsubsection{Bucketing Strategy}
To achieve efficient retrieval within the large-scale unstructured question bank $\mathcal{Q}$, we first perform a structural reorganization. In the memory module of the Selection Agent, drawing inspiration from database indexing principles \cite{douze2025faiss}, we propose an IRT parameter stratification mechanism to construct a multi-dimensional index structure. Specifically, for each knowledge concept $k \in \mathcal{K}$, we partition the questions into distinct ‘‘ buckets ’’ based on the question difficulty parameter $b_j$ and discrimination parameter $a_j$ of question $q_j$. The bucket $\mathcal{B}^{k}_{l,d}$ corresponding to knowledge concept $k$ is defined as follows, and the question $q_j$ with this knowledge concept is stored in it:
\begin{align}
    \mathcal{B}_{l,d}^k = \{ q_j \in \mathcal{Q} \mid \text{KC}(q_j) = k, \text{Dif}(b_j) = l, \text{Dis}(a_j) = d \},
\end{align}
where KC, Dif, and Dis represent the knowledge concept, difficulty, and discrimination features of question $q_j$, respectively. We map the numerical distribution interval of $b_j$ into three difficulty levels $l \in \{\text{Easy, Medium, Hard}\}$, and partition the numerical distribution of $a_j$ into two discrimination levels $d \in \{\text{Low, High}\}$. This bucketing structure significantly reduces the question search space from a linear global complexity of $O(|\mathcal{Q}|)$ to a bucket-level complexity of $O(|\mathcal{B}_{avg}|)$ (where $|\mathcal{B}_{avg}|$ denotes the average number of questions per bucket). This not only substantially enhances retrieval real-time performance but also lays the data foundation for the subsequent Agent's “ Coarse-to-Fine ”  hierarchical decision-making.

\subsubsection{Coarse-to-Fine}
Based on the constructed multi-dimensional bucket index, the decision process of the Selection Agent is decoupled into two phases: Coarse-grained Instructional Planning and Fine-grained Hierarchical Retrieval.

\textbf{Phase 1: Coarse-grained Instructional Planning.} In this phase, the Selection Agent does not directly process specific question entities but acts as an Instructional Planner, making macro-decisions at the knowledge concept level. Leveraging the LLM's powerful semantic reasoning capabilities and its understanding of the  $\mathcal{G}$, the agent infers the next most valuable target knowledge concept ($k_{tgt}$) based on the examinee's historical response  $\mathcal{H}_{t-1}$ and current latent ability state $\hat{\theta}^{(t-1)}$. This decision process mainly follows two pedagogical principles:
(1) Prerequisite Dependency: If the examinee fails at the current concept, the LLMs traces back along the $\mathcal{G}$ to select prerequisite concepts for remedial testing.
(2) Zone of Proximal Development (ZPD)\cite{wass2014sharpening}: If the examinee performs well, the LLMs recommends successor or highly related advanced concepts.
The output of this phase is an explicit retrieval instruction $k_{tgt}$. The system then automatically maps the target  bucket  $\mathcal{B}_{tgt}$.

\textbf{Phase 2: Fine-grained Hierarchical Retrieval Strategy.} Due to the inherent sparsity in the distribution of knowledge concepts and difficulty within real-world question banks, retrieving directly from the target bucket $\mathcal{B}_{tgt}$ poses a high risk of deadlock—either running out of suitable questions or facing severe difficulty mismatches. To address this, we designed a Four-Level Stratified Retrieval Strategy. This strategy adheres to the principle of Prioritizing Pedagogical Coherence with Difficulty Adaptation as a Safety Net, locking onto the optimal question $q_t$ by progressively relaxing constraints. First, we define a comprehensive scoring function $S(q)$ to evaluate the compatibility of a candidate question $q$ with the current state:
\begin{align}
    S(q) = w_1 \cdot Gap(q) + w_2 \cdot \mathcal{C} \cdot \mathbb{I}(k_q = k_{tgt}) + w_3 \cdot (1 - \text{Prof}_q),
\end{align}
where the first term penalizes the difficulty deviation $Gap(q) = |b_q - \theta_{t-1}|$. The second term rewards knowledge concept hits, where $k_q$ denotes the concept of the candidate question. Here, $\mathcal{C}$ represents the Planning Confidence of the Selection Agent, reflecting its certainty regarding the current target concept. This mechanism acts as a soft-gate, automatically relaxing semantic constraints when the decision is uncertain to prioritize difficulty adaptation while attempting to maintain pedagogical coherence. The third term serves as the Weakness Priority factor, where $\text{Prof}_q$ indicates the examinee's historical proficiency on the concept of question $q$. This term utilizes $(1 - \text{Prof}_q)$ to assign higher weights to concepts with lower mastery, effectively implementing a remedial strategy to explore the examinee's cognitive weak concepts. The specific retrieval process within a bucket proceeds as follows:

\begin{itemize}
    \item \textbf{Tier 1: Strict Target Matching.} The system prioritizes retrieval within the target difficulty bucket $B_{tgt}$ of the target knowledge concept. If the difficulty gap of the optimal question within the bucket satisfies the strict threshold constraint (i.e., $\text{Gap}(q) < \epsilon_{strict}$), this question is directly selected. This achieves optimal difficulty adaptation while simultaneously maintaining instructional continuity.
    \item \textbf{Tier 2: Intra-Concept Relaxation.} If Tier 1 retrieval fails (i.e., the bucket is empty or the gap is too large), the system expands the retrieval scope to \textit{all} difficulty buckets under $k_{tgt}$. If the optimal question found within this scope satisfies the relaxed threshold constraint (i.e., $\text{Gap}(q) < \epsilon_{relaxed}$, where $\epsilon_{relaxed} > \epsilon_{strict}$), it is selected. This strategy prioritizes maintaining the instructional continuity of the knowledge concept by sacrificing a degree of difficulty adaptation precision.
    \item \textbf{Tier 3: Neighbor Topological Hop.} If Tier 2 still fails to meet the conditions, it indicates a severe mismatch between available questions for this concept and the examinee's current ability. At this concept, the system searches the topological neighbor nodes (neighbor concepts) of $k_{tgt}$ along the  $\mathcal{G}$\cite{huang2021knowledge}. If a question in the neighbor nodes offers significantly better difficulty adaptation (i.e., $\text{Gap}_{neighbor} \ll \text{Gap}_{tier2}$), the system performs a semantic hop. This strategy effectively resolves the single-concept question exhaustion issue by leveraging graph connectivity.
    \item \textbf{Tier 4: Global Robust Fallback.} Serving as the final guarantee of system robustness, if no question satisfying the maximum tolerance ($\tau_{max}$) is found within the aforementioned scopes, the system performs a global search across the entire question bank $\mathcal{Q} \setminus \mathcal{H}_{t-1}$. At this stage, the system selects the mathematically best-matched question based solely on the comprehensive scoring formula $S(q)$, ensuring the testing process does not suffer interruption.
\end{itemize}

Through this cascading mechanism, AgentCAT fully leverages the instructional planning capabilities of LLMs while simultaneously guaranteeing the adaptive convergence of CAT via rigorous mathematical constraints.

\subsection{Supervisor Module}
Due to the hallucination issues and generative uncertainty inherent in LLMs\cite{zhang2025llm}, directly adopting the outputs of the Examinee Agent may lead to drift or oscillation in the CAT system's ability estimation\cite{shinn2023reflexion}. Acting as the ‘‘ Auditor ’’ and ‘‘  Arbitrator ’’ of the entire simulation system, the Supervisor Module $\mathcal{V}$ is designed to ensure the stability and convergence of the assessment process through dual-side output auditing and robust ability update mechanisms\cite{ma2025advancing}.

\subsubsection{Dual-Agent Auditing Mechanism}
Before executing the ability update, the Supervisor Module $\mathcal{V}$ constructs two "safety guardrails" to intercept and audit the outputs of the Selection Agent and the Examinee Agent in real-time.

\textbf{Validity Auditing for Selection Agent (Selection Auditing)}:
To prevent the selection strategy from falling into local loops or recommending invalid questions, we design an IRT-probability-based validity check. For a selected question $q_t$, the system calculates the predicted correct response probability $P(\hat{\theta}^{(t-1)})$ under the current ability $\hat{\theta}^{(t-1)}$. If $P(\hat{\theta}^{(t-1)})$ falls outside the valid interval, it indicates the question difficulty is too extreme. Such questions, due to their Fisher Information approaching zero, not only fail to provide effective diagnostic gain but may also lead to examinee frustration or ineffective practice \cite{lord2012applications}. Consequently, such selections are Rejected, triggering the Selection Agent's re-planning mechanism. Simultaneously, the system maintains a short-term question queue $\mathcal{Q}_{recent}$ to forcibly intercept recurring questions (i.e., $q_t \notin \mathcal{Q}_{recent}$), strictly ensuring the diversity of the test sequence.

\textbf{Cognitive Auditing for Examinee Agent (Examinee Auditing)}: To detect whether the Examinee Agent is trapped in a reasoning loop, we introduced a text detection mechanism based on Jaccard similarity \cite{nguyen2025enhancing}. Let $T^{(t)}_{idea}$ denote the reasoning text generated by the Examinee Agent at step $t$. The system calculates the bag-of-words similarity between it and the most recent historical reasoning $T^{(t-1)}_{idea}$:
\begin{align}
    J(T^{(t)}{idea}, T^{(t-1)}{idea}) = \frac{|\text{Set}(T^{(t)}{idea}) \cap \text{Set}(T^{(t-1)}{idea})|}{|\text{Set}(T^{(t)}{idea}) \cup \text{Set}(T^{(t-1)}{idea})|},
\end{align}
where $\text{Set}(TE)$ represents the set of unique tokens contained in text $TE$. If $J > 0.90$, the system judges the response as an invalid repetition, triggering a regeneration or truncation mechanism to prevent erroneous cognitive states from self-reinforcing in a closed loop.

\subsubsection{Robust Ability Update}
Upon obtaining audited responses, the system faces a core challenge: how to reconcile the simulated response $y_{agent}$ from the Examinee Agent with the historical ground truth label $y_{truth}$ (if it exists). To balance between preventing ability inflation and avoiding ability underestimation, we propose an Asymmetric Soft-Overwriting Strategy. We define $w_t$ as the examinee agent's accumulated confidence. The determination logic for the final label $y_t$ is as follows:

\begin{itemize}
    \item \textbf{Consistency Scenario:} If $y_{agent} = y_{truth}$, the system directly adopts it and increments  $w_t$.
    
    \item \textbf{Conflict Scenario A: Preventing Ability Inflation.} When $y_{agent} = 1$ but $y_{truth} = 0$, there is a high risk of false positives. Directly trusting the Agent would cause the ability estimate to skyrocket abnormally. Therefore, we set an extremely strict overwriting threshold: we allow adopting $y_{agent}=1$ only if $w_t$ is sufficiently large \and the IRT model also predicts a high probability of correctness; otherwise, $y_{truth}=0$ is forcibly maintained as a safety baseline. This ensures the system prefers conservative estimation over unreliable score inflation.
    
    \item \textbf{Conflict Scenario B: Preventing Ability Underestimation.} When $y_{agent} = 0$ but $y_{truth} = 1$, directly trusting the Agent would lead to erroneous underestimation. Typically, we prioritize the historical truth $y_{truth}=1$, assuming the examinee has mastered the concept. \textbf{Exception Handling}: However, to capture the dynamic forgetting effect, if strong evidence exists---i.e., the agent is highly credible and the IRT model predicts a very low probability of correctness for the current ability---we allow the system to overwrite the historical truth and adopt $y_{agent}=0$. This enables AgentCAT to keenly diagnose knowledge blind spots that were once mastered but are now forgotten.
    
    \item \textbf{Counterfactual Scenario:} If $y_{truth} = \text{None}$, $y_{agent}$ is fully adopted. This embodies the core value of AgentCAT, leveraging the cognitive simulation capabilities of the Examinee Agent to fill the gaps caused by data sparsity and explore unknown knowledge domains.
\end{itemize}

Through this asymmetric mechanism, the Supervisor constructs a ‘‘ ceiling ’’ to prevent ability inflation and a ‘‘ floor ’’ to prevent unreasonable underestimation, while retaining the flexibility to capture forgetting phenomena.

Upon acquiring the final label $y_t$, we adopt the principles of the 2PL-IRT model to update the estimated ability value $\hat{\theta}^{(t)}$. In traditional CAT research, ability estimation modules typically rely on CDM that have been pre-trained on massive historical interaction datasets~\cite{zhuang2026survey, ma2025diffusion,wang2025explicit}. However, in the full-process simulation scenario of AgentCAT, the Supervisor module cannot rely on pre-learned examinee embedding vectors. Instead, it must initiate from an assessment starting state ($\hat{\theta}^{(0)}$) and achieve rapid convergence of the ability estimation process relying solely on the real-time simulated interaction stream. Therefore, addressing this unique challenge, we propose an Online Robust Gradient Descent Strategy. This strategy models ability assessment as a dynamic parameter tracking process, incorporating physical constraints to accommodate the generative characteristics of LLMs.

To achieve rapid convergence and robust estimation of ability values within a single-turn agent interaction stream, we model the ability update as an online gradient descent process characterized by an adaptive step size and gradient clipping. The update formula for the ability value $\hat{\theta}^{(t)}$ at step $t$ is defined as follows:
\begin{align}
    \hat{\theta}^{(t)} = \hat{\theta}^{(t-1)} + \eta_{eff} \cdot \Psi(\nabla_{total}),
\end{align}
where the total gradient $\nabla_{total}$ is a linear combination of the data likelihood gradient $\nabla_{data}$ and the prior regularization gradient $\nabla_{reg}$:

\begin{align}
    \nabla_{total} = \underbrace{a_t (y_t -P(\hat{\theta}^{(t-1)})}_{\nabla_{data}}) + \underbrace{-\lambda_{reg} (\hat{\theta}^{(t-1)} - \hat{\theta}^{(0)})}_{\nabla_{reg}}.
\end{align}

$a_t$ denotes the discrimination parameter of the current question $q_t$, and $\eta_{eff}$ is the effective learning rate, controlling the magnitude of the single-step update. $\Psi(\cdot)$ represents the saturation damping function, designed to suppress gradient explosion. $P(\hat{\theta}^{(t-1)})$ is the predicted probability of a correct response. $\lambda_{reg}$ is the regularization intensity coefficient, preventing the estimated value from deviating excessively from the initial estimate under data sparsity. Specifically, $\nabla_{data}$ is derived from the log-likelihood derivative of the 2PL-IRT model, reflecting the direct corrective power of the current response $y_t$ on the ability estimate: when a examinee answers correctly despite a low predicted probability, a positive gradient is generated to push $\hat{\theta}^{(t)}$ upward, and vice versa. Meanwhile, $\nabla_{reg}$ originates from a Gaussian prior constraint, acting as an elastic restoring force pointing towards the initial ability $\hat{\theta}^{(0)}$. This prevents drastic, non-physical parameter drift caused by data sparsity or noise interference during the early stages of simulation. To address the issues of overconfidence and information fluctuation in generated data\cite{yao2024adard}, we introduce robust control mechanisms based on the gradients described above:

(1) Saturation Damping ($\Psi$). This mechanism applies to the total gradient and is designed to mitigate the gradient explosion problem caused by the Agent's hallucinated certainty\cite{kim2025splitnet,yazdani2023robust}. In traditional SGD, consecutive correct responses lead to the continuous accumulation of positive gradients. However, in simulation, the Examinee Agent might continue to output correct labels even when the concept is already fully mastered (i.e., predicted probability $P(\hat{\theta}^{(t-1)}) \to 1$). At this stage, the marginal information gain diminishes. We define a damping function $\Psi(g)$ to forcibly attenuate the gradient magnitude when the model enters a saturation zone of extremely high predictive probability:
\begin{equation}
    \Psi(g) = 
    \begin{cases} 
    \gamma \cdot g, & \begin{aligned} 
                        \text{if } & (y_t=1 \land P(\hat{\theta}^{(t-1)}) > \tau_{sat}) 
                                   & \lor (y_t=0 \land P(\hat{\theta}^{(t-1)}) < 1 - \tau_{sat}) 
                      \end{aligned} \\
    g, & \text{otherwise}
    \end{cases}
\end{equation}
where $\gamma \in (0, 1)$ is the damping coefficient, and $\tau_{sat}$ is the saturation threshold used to delineate low-information regions. When the model's predicted probability $P(\hat{\theta}^{(t-1)})$ falls outside the interval $[1 - \tau_{sat}, \tau_{sat}]$, it indicates that the current question is too extreme  relative to the examinee's ability, or that the Examinee Agent is exhibiting overconfidence. In such cases, the system determines that it has entered a saturation state and triggers gradient damping to prevent parameter drift. This simulates the plateau effect in learning curves, effectively preventing the artificial inflation of ability values.

(2) Fisher Information-based Adaptive Learning Rate ($\eta_{eff}$). This mechanism operates on the update step size, dynamically adjusting the magnitude of updates based on question quality. We utilize Fisher Information to quantify the contribution of the current question $q_t$ to the precision of ability estimation. Specifically, defining $I(\theta)$ as the Fisher information of the current question at ability level $\theta$, we dynamically scale the base learning rate $\eta_{base}$ as follows:
\begin{align}
    \eta_{eff} = \eta_{base} \cdot \left( \beta + (1-\beta) \cdot \frac{I(\theta)}{I_{max}} \right),
\end{align}
where $I(\theta)$ represents the instantaneous Fisher information provided by question $q_t$ at the previous ability estimate $\hat{\theta}^{(t-1)}$. $I_{max}$ denotes the Global Maximum Information Constant (i.e., the theoretical maximum information yieldable by any question in the bank), serving to normalize $I(\theta)$ into the $[0, 1]$ interval. $\beta \in (0, 1)$ is the base retention coefficient, ensuring that the system maintains a minimum update step size even in low-information regions. This mechanism endows the system with a weighted update capability: when encountering high-efficacy questions (characterized by high discrimination and moderate difficulty, where $I(\theta)$ approaches $I_{max}$), the system deems the current gradient direction highly reliable, thereby amplifying $\eta_{eff}$ to accelerate convergence. Conversely, when facing ineffective questions with low discrimination or extreme difficulty, the system automatically attenuates the update magnitude to suppress noise interference.

\section{Experiments}

To comprehensively evaluate the effectiveness of the AgentCAT framework, this section aims to address the following four core research questions (RQs):
\begin{itemize}
    \item ~\textbf{RQ1:} Can AgentCAT accurately estimate examinee ability comparable to established CAT methods?
    \item ~\textbf{RQ2:} How effective is the proposed Coarse-to-Fine selection mechanism in balancing difficulty adaptation with information gain?
    \item ~\textbf{RQ3:} How does the testing path generated by AgentCAT perform in terms of consistency and interpretability over the knowledge concept topology?
    \item ~\textbf{RQ4:} Does the AgentCAT framework demonstrate robust ability estimation capabilities in counterfactual scenarios where ground-truth labels are absent, by leveraging the Examinee Agent simulation?
\end{itemize}

\subsection{Experimental Settings}

\subsubsection{Dataset Description}

In this study, we utilize two datasets: XES3G5M and C\_797404, are described in detail in the table \ref{dataset}. XES3G5M is a large-scale educational interaction dataset collected from a real-world online mathematics learning platform\cite{liu2023xes3g5m}. C\_797404 is a course dataset regarding wine knowledge from the MOOCradar\cite{yu2023moocradar}. Both datasets are equipped with rich auxiliary information, including the Knowledge components  corresponding to each question, question types, textual content,  and response timestamps. This information far exceeds traditional datasets that rely solely on question IDs and response labels, facilitating the construction of more fine-grained and context-aware learning behavior models. Following the methodology of Agent4Edu\cite{gao2025agent4edu}, we split the dataset into training and testing sets. From the training set, we extract information such as the examinee's recently frequented knowledge concepts and historical accuracy rates to construct the Examinee Agent's Profile. We treat the ability values obtained by training an IRT model on the complete interaction records as the examinee's ground truth ability. Meanwhile, the ability values obtained by training solely on the training set serve as the Examinee Agent's Initial Ability. The CAT assessment process is conducted on the testing set.
\begin{table}[!t]
  \centering
  \caption{The statistics of the datasets.}
  \vspace{1pt}
  \renewcommand{\arraystretch}{1.3} 
  \setlength{\tabcolsep}{25pt}      
  \begin{tabular}{lrr}             
    \toprule
    DataSet    & XES3G5M   & C\_797404 \\
    \midrule
    \#Examinee & 18,066    & 1,221    \\
    \#Question & 7,652     & 184      \\
    \#Concept  & 1,175     & 253      \\
    \#Log      & 5,549,635 & 174,256  \\
    \bottomrule
  \end{tabular}
  \label{dataset}
  \vspace{-5mm}
\end{table}
\subsubsection{Baseline approaches}
To validate the effectiveness and rationality of the AgentCAT simulation framework in CAT research, we selected several classic question selection strategies as baselines. During the experiments, to ensure a fair comparison and control variables, we maintained the overall interaction architecture of AgentCAT unchanged, solely replacing the Selection Agent with traditional CAT selection algorithms. AgentCAT is designed to simulate realistic online assessment processes, emphasizing decision-making based on the agent's reasoning capabilities in cold-start or real-time interaction scenarios devoid of accumulated historical data. Therefore, selecting rule-based algorithms that are similarly training-free and plug-and-play as baselines better highlights the fundamental difference between LLM-based reasoning and traditional mathematical logic in handling assessment protocols. Based on these considerations, we selected the following four representative methods:~(1)~Random: It randomly selects a question from the question bank. (2)~KLI\cite{chang1996global}: It employs the global metric of KL information to balance the local precision and global discrimination of question selection. (3)~MFI\cite{lord2012applications}: It calculates the Fisher Information for all candidate questions under the current ability estimate  and selects the question with the maximum information. (4)~MAAT\cite{bi2020quality}: It is a selection method based on active learning, choosing the question that results in the maximum expected change in ability.

\subsubsection{Large Language Model}
In our experiments, we categorized the backbone LLMs into two types: API-based and locally deployed. Regarding API-based models, we selected GLM-4-FlashX and GLM-4.7-FlashX from the Zhipu AI GLM series. The former is a high-speed, low-cost model characterized by extreme cost-efficiency and ultra-fast inference speed, with native support for tool calling; while the latter serves as a lightweight high-speed flagship model, featuring robust capabilities in programming and tool usage, making it highly suitable for agent-based tasks. For locally deployed models, we utilized the Qwen series: Qwen2.5-7B and Qwen3-4B. The former is a lightweight general-purpose model with a 7B parameter scale, effectively balancing performance with efficiency to adapt to various tasks. The latter is an efficient inference model with 4B parameters, supporting Chain-of-Thought (CoT) switching and tool calling, striking a balance between generation quality and resource consumption. Additionally, for the implementation of baseline methods, we consistently employed GLM-4-FlashX as the base model to drive the Examinee Agent.

\subsubsection{Prompt settings}
Within the multi-agent collaborative framework of AgentCAT, this study employs a structured prompt template dynamically assembled through code logic. Taking the selection agent as an example, its comprehensive prompt is constructed from three core modules:~(1) Explicitly endowing the model with the identity of a selection agent, alongside strict formatting instructions for its output.~(2) Dynamically injecting the current learning state of the examinee, encompassing the current ability estimate, knowledge concept proficiency, and the system's current instructional target.~(3) Injecting the response performance on the previous exercise, including the response outcome and knowledge concept topological dependencies, while presenting a summarized display of the currently available exercise pool, which is clustered into six specific feature buckets based on difficulty and discrimination. The core prompt template for selecting an agent is shown in the figure~\ref{selection_agent_prompt}.

\begin{figure}[htbp]
\centering
\begin{tcolorbox}[
    enhanced,
    width=\columnwidth, 
    colback=gray!2!white,
    colframe=gray!75!black,
    title=\textbf{Prompt Template for Selection Agent},
    fontupper=\footnotesize\ttfamily,
    halign=flush left,
    arc=2mm,
    boxrule=0.5pt,
    left=2mm,
    right=2mm,
    top=1mm,
    bottom=1mm
]

\textbf{[System Instruction]} \\
You are a question selection agent, familiar with computerized adaptive testing principles, able to select appropriate questions based on student answering situations. \\
Please use english for all explanatory text; but must strictly use the english tags and bucket names given below for structured output: \\
1) Output tags: 'Knowledge:' and 'Bucket:' \\
2) Bucket names (choose one): \textcolor{blue}{\{bucket\_options\}} \\
3) Please copy the knowledge point name exactly from the 'Available Buckets Summary' below. \\
Final output must only include the two lines below.

\vspace{0.4em}
\tcbline
\vspace{0.4em}

\textbf{[Student Profile]} \\
\textcolor{blue}{\{student\_profile\_data\}} 

\vspace{0.5em}
\textbf{[Recent Answer Result]} \\
- Knowledge Point: \textcolor{blue}{\{current\_kc\}} \\
- KC Reliability: \textcolor{blue}{\{kc\_reliability\}} \\
- Answer: \textcolor{blue}{\{student\_answer\}} \\
- Score: \textcolor{blue}{\{score\}} \\
- Corrective Feedback: \textcolor{blue}{\{feedback\}} \\
- Knowledge Dependencies: \textcolor{blue}{\{dependencies\_str\}}

\vspace{0.8em}
\textbf{[Available Buckets Summary]} \\
\textcolor{blue}{\{stats\_text\}}

\vspace{0.8em}
\tcbline
\vspace{0.8em}

\textbf{[Instructions]} \\
Select the most appropriate knowledge point and bucket based on student ability, knowledge proficiency, dependency relationships, and availability. 
If KC Reliability is low ($<0.35$), rely less on the last predicted KC and prioritize ability/difficulty matching. Do not output specific exercise IDs.

\vspace{0.8em}
\textbf{[Output Format]} \\
Knowledge: \textless knowledge\_name\textgreater \\
Bucket: \textless bucket\_name\textgreater

\end{tcolorbox}
\caption{Structured Prompt Template for the Exercise Selection Agent in AgentCAT}
\label{selection_agent_prompt}
\end{figure}

\subsubsection{Parameter Settings}
We defined the following hyperparameters in our experiments: In the Supervisor module, we initialized the examinee ability $\hat{\theta}^{(0)} = 0.0$ with a Gaussian prior. The hyperparameters for the Online Robust Gradient Descent were set as follows: base learning rate $\eta_{base} = 0.05$ with a decay factor of $0.98$, regularization coefficient $\lambda_{reg} = 0.02$, and gradient clipping threshold set to $1.0$. To mitigate the plateau effect, the saturation threshold $\tau_{sat}$ was set to $0.85$ with a damping factor $\gamma = 0.35$. For the Selection Agent, the weights in the scoring function $S(q)$ were empirically set to $w_1 = 5.0$ (difficulty adaptation), $w_2 = 0.5$ (knowledge coherence), and $w_3 = 0.05$ (weakness priority), reflecting a design philosophy that prioritizes difficulty matching as the primary constraint. In the label fusion stage, we employed asymmetric thresholds for the soft override policy: a stricter confidence threshold ($w_t \ge 0.75$) was applied to prevent false positive overestimation, while a moderate threshold ($w_t \ge 0.60$) was used for false negatives. The thresholds for the four-tier retrieval strategy were empirically set as follows: $\epsilon_{strict} = 0.25$ for the ideal zone, $\epsilon_{relaxed} = 0.40$ for the compromise zone, and $\tau_{max} = 0.60$ as the maximum tolerance before triggering the global fallback search.

\subsection{Ability Estimation(RQ1)}
\begin{figure}[!ht]
    \centering
    
    \subfloat[Performance on XES3G5M]{
        \includegraphics[width=0.48\linewidth]{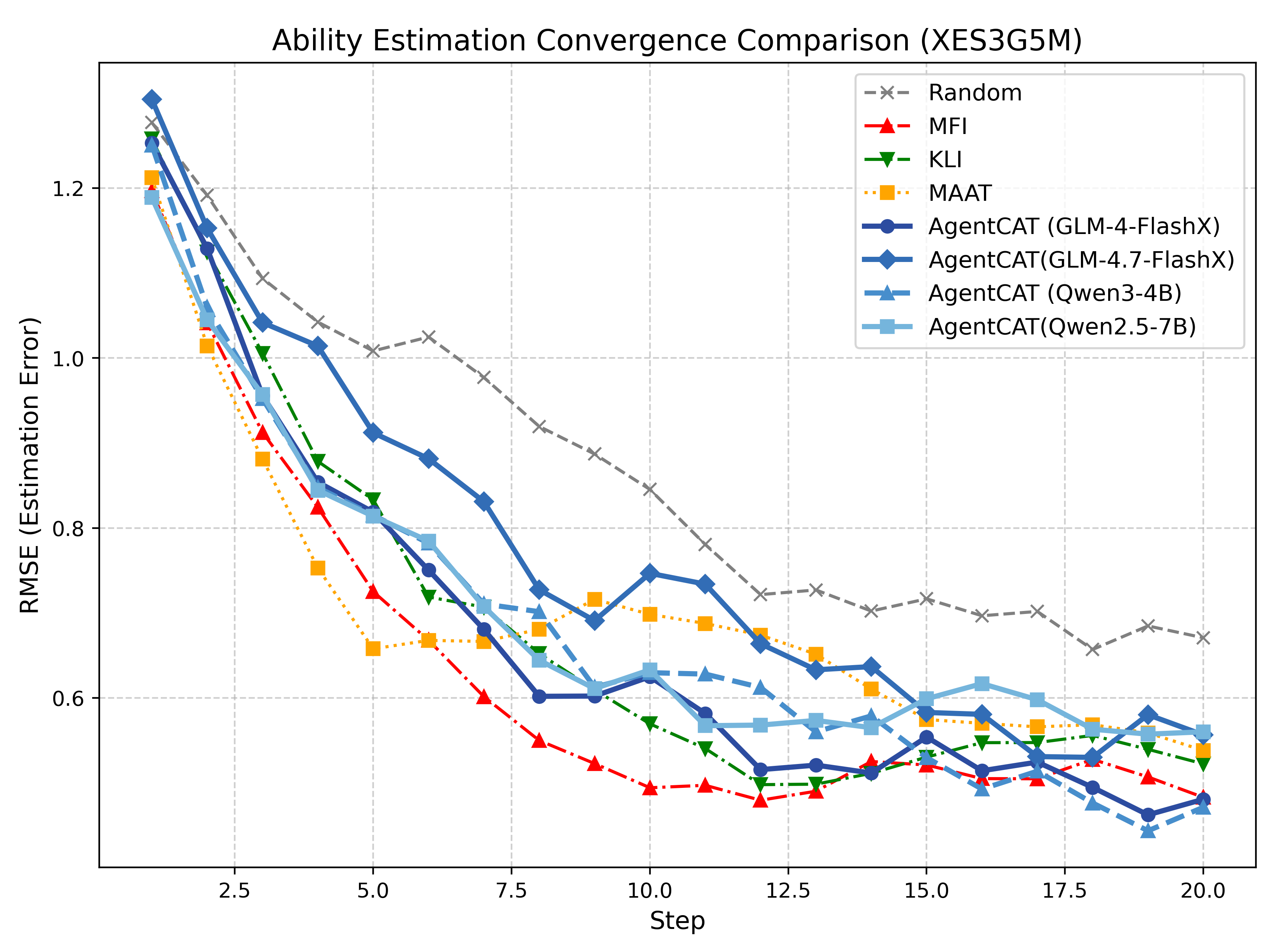}
        \label{fig:rmse_xes}
    }
    \hfill 
    \subfloat[Performance on C\_797404]{
        \includegraphics[width=0.48\linewidth]{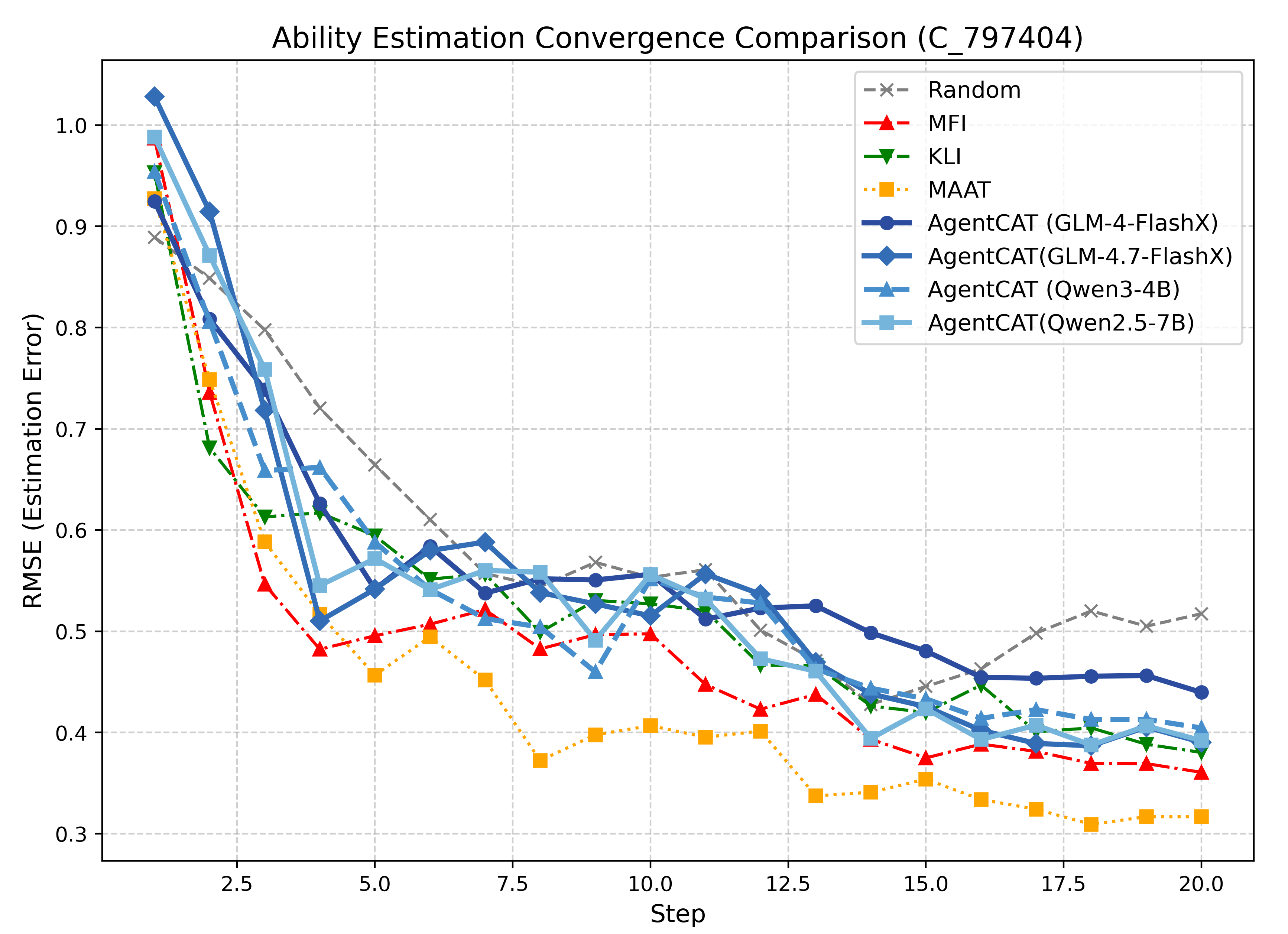}
        \label{fig:rmse_mooc}
    }
    
    \caption{Ability estimation convergence comparison (RMSE) on two datasets. AgentCAT demonstrates consistent convergence trends comparable to classical mathematical baselines across different LLMs backbones.}
    \label{fig:rmse_convergence}
\end{figure}

This section aims to validate the effectiveness, robustness, and generalizability of AgentCAT as a high-quality proficiency assessment simulation system in evaluating the examinee's ability under a multi-agent interactive framework. We adopt an "Ability Benchmark Alignment" scheme, utilizing the IRT parameters trained on the full dataset as the examinees' ground-truth abilities, and initializing the examinee agent with the training set. The experiment involves stratified random sampling of 30 examinees across different ability ranges from both datasets, evaluating the simulation fidelity by observing the Root Mean Square Error (RMSE)~\cite{wang1998properties} variations between the system's estimated values and the ground-truth values. As illustrated in Figure \ref{fig:rmse_convergence}, the experimental results on both datasets reveal three highly consistent core conclusions:~(1) Significance of Ability Estimation Convergence: Whether on XES3G5M or C\_797404, AgentCAT's RMSE curves exhibit a monotonically decreasing trend that aligns with the principles of educational assessment. This consistent cross-dataset performance demonstrates that the "coarse-to-fine" strategy of the selection agent and the auditing mechanism of the supervisor do not overfit to a single dataset; rather, they can effectively overcome the generative noise of LLMs and stably extract diagnostic information from natural language interactions across diverse domains.~(2) Robustness of the Educational Simulation Mechanism: In both experimental scenarios, AgentCAT driven by different LLM backbones (e.g., GLM-4, Qwen2.5) demonstrates highly convergent slopes and error levels. This further indicates that the framework's effectiveness stems from its built-in pedagogical constraints and the robust design of the multi-agent collaborative mechanism, rather than relying on the reasoning capabilities of specific models. This provides a solid reliability guarantee for deploying models of varying specifications within this simulation system in the future.~(3) Statistical Reliability: In the later stages of testing (steps 15-20), AgentCAT's error level is on par with classic mathematical optimization strategies such as MFI and KLI. This result compellingly proves that natural language-driven agent simulation systems have achieved a level of statistical reliability in ability estimation equivalent to that of rigorous mathematical models.

\vspace{-0.5em}
\subsection{Selection Balance and Information Content of Exercises(RQ2)}
This section delves into the micro-level interaction behaviors of AgentCAT, aiming to verify that its ability estimation convergence is not a result of random walks, but stems from high-quality adaptive selection. The base LLMs used in this experiment is GLM-4-FlashX.

\begin{figure}[htbp]
    \centering
    
    \subfloat[Difficulty Matching Error ($|b_t - \hat{\theta}^{(t-1)}|$)]{
        \includegraphics[width=0.48\linewidth]{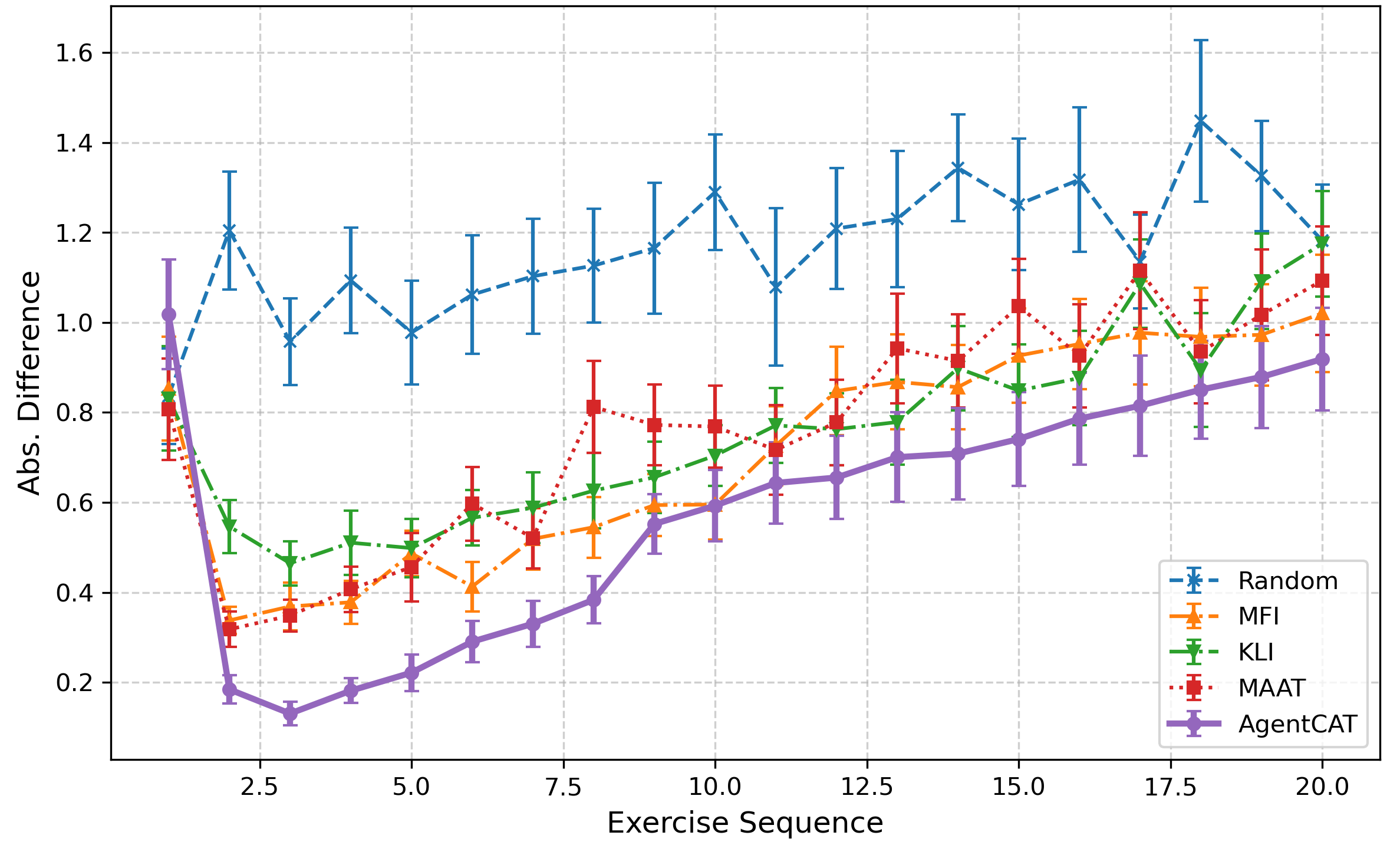}
        \label{fig:diff_gap_XES3G5M}
    }
    \hfill 
    \subfloat[Average Fisher Information ($I(\theta)$)]{
        \includegraphics[width=0.48\linewidth]{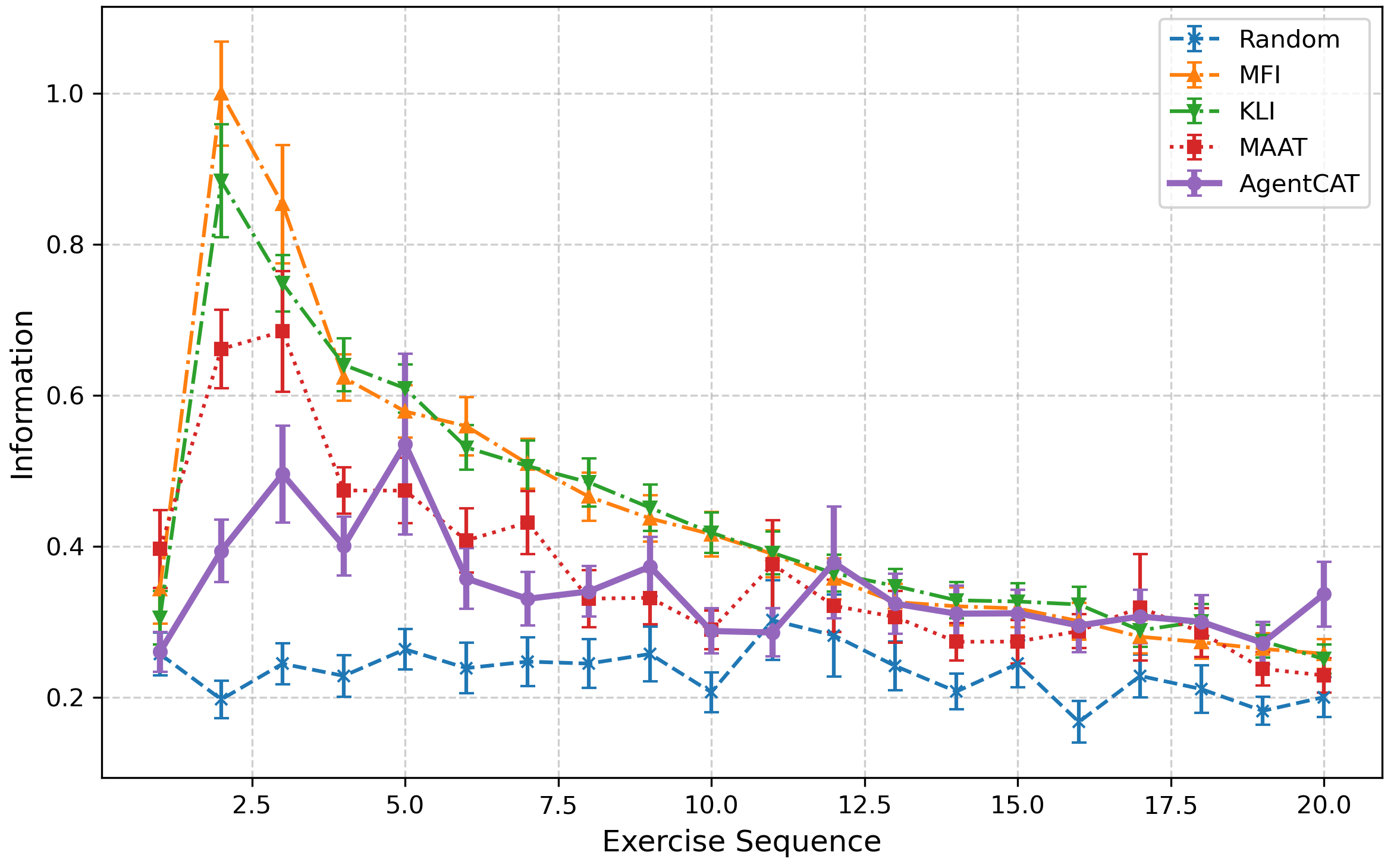}
        \label{fig:fisher_info_XES3G5M}
    }
    
    \caption{Micro-level interaction analysis on XES3G5M.}
    \label{fig:exp2_analysis_XES3G5M}
\end{figure}

\begin{figure}[htbp]
    \centering
    
    \subfloat[Difficulty Matching Error ($|b_t - \hat{\theta}^{(t-1)}|$)]{
        \includegraphics[width=0.48\linewidth]{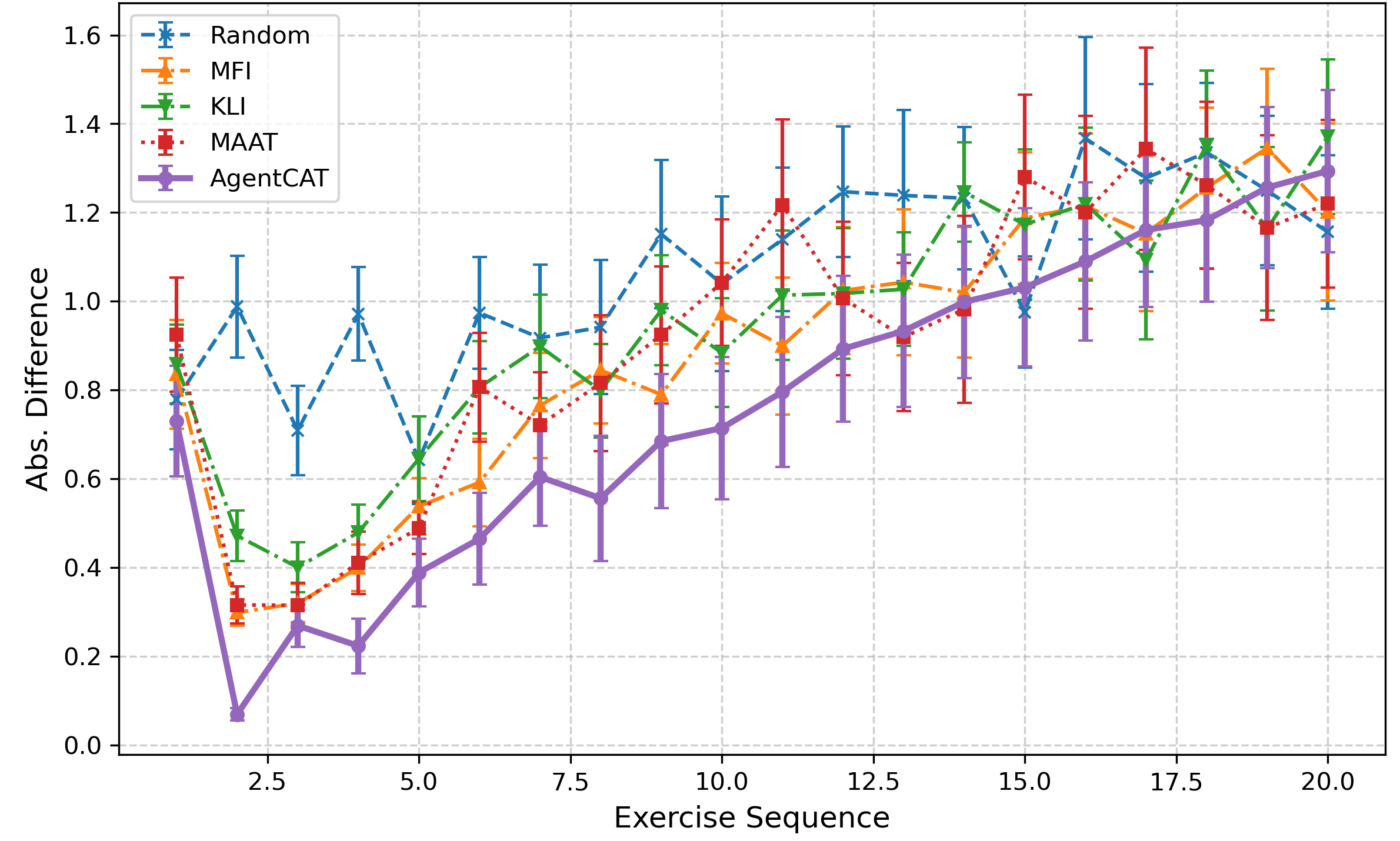}
        \label{fig:diff_gap_C_797404}
    }
    \hfill 
    \subfloat[Average Fisher Information ($I(\theta)$)]{
        \includegraphics[width=0.48\linewidth]{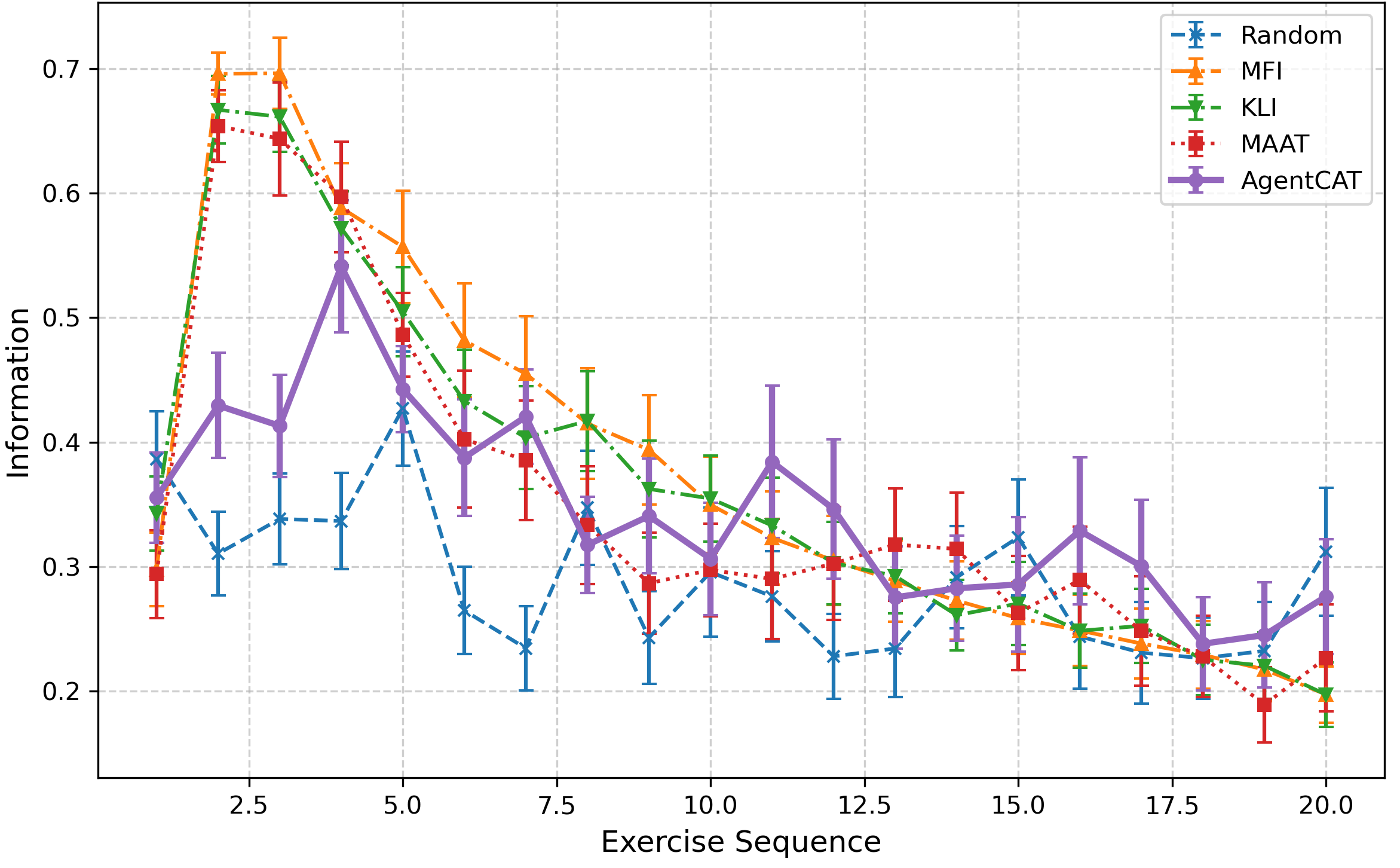}
        \label{fig:fisher_info_C_797404}
    }
    
    \caption{Micro-level interaction analysis on C\_797404.}
    \label{fig:exp2_analysis_C_797404}
\end{figure}

\textbf{Difficulty Match Analysis.} Authentic educational assessments require that the exercise difficulty ($b_t$) be dynamically anchored to the examinee's ability ($\hat{\theta}^{(t-1)}$), ensuring that the test remains challenging without inducing frustration. As illustrated in Figures \ref{fig:diff_gap_XES3G5M}and \ref{fig:diff_gap_C_797404}, the experimental results across both datasets reveal highly consistent adaptive characteristics:~(1) Rapid Positioning during the Cold-Start Phase: In the initial testing phase (Q1-Q5), AgentCAT achieves a steep decline in difficulty deviation ($|b - \hat{\theta}|$) across both datasets. Taking XES3G5M as an example, its deviation converges rapidly from 1.018 to 0.221. This rapid cross-dataset convergence confirms that the "coarse-to-fine" bucketing strategy of the selection agent can effectively overcome the cold-start challenge, swiftly locking onto the ZPD of examinees across different domains.~(2) Balance of Exploration and Exploitation: In the intermediate testing phase (Q6-Q15), AgentCAT's difficulty deviation stabilizes within a low-value range, which is significantly lower than that of baseline methods pursuing extreme mathematical matching. The root of this discrepancy lies in the varying quality of data supplied to the supervisor by different selection strategies. For instance, although both share the identical ability estimator, MFI adopts a strict numerical anchoring strategy. Once an examinee's ability estimate plummets due to early-stage noise, MFI falls into a low-difficulty trap: it continuously pushes low-difficulty exercises. Since the information gain from high-ability examinees correctly answering low-difficulty exercises is extremely marginal, it is difficult for the supervisor to obtain sufficient gradients from these responses to rapidly correct the ability estimate, causing the system to fall into prolonged estimation bias. In contrast, AgentCAT's semantic exploration mechanism (such as the Tier 3 topological neighbor exploration) behaves more like an experienced teacher. Upon detecting that the assessment has stagnated, it proactively introduces appropriate difficulty fluctuations and cross-concept probing, thereby rapidly acquiring more discriminative authentic feedback, which vastly enhances the pedagogical resilience and error-correction capability of the system.

    
    
    
    

\textbf{Information Gain Analysis.} Fisher Information quantifies the contribution of an questions to the precision of ability estimation. As illustrated in Figure \ref{fig:fisher_info_XES3G5M} and \ref{fig:fisher_info_C_797404}:~(1) Avoiding Extreme Exercise Selection: Taking XES3G5M as an example, MFI and KLI tend to select extreme exercises with high discrimination (where the information peak approaches 1.0). Such idiosyncratic or overly difficult exercises can trigger severe cognitive overload and anxiety in examinees. In contrast, AgentCAT maintains its average information gain within a robust range of 0.3 to 0.5. It proactively abandons the greedy pursuit of extreme statistical indicators, favoring moderate exercises that both possess diagnostic value and accommodate the examinee's cognitive load.~(2) Unification of Semantics and Mathematics: As the test progresses (post-Q10), AgentCAT's information trajectory gradually converges to approximately 0.3, highly aligning with the late-stage performance of MFI. This compellingly demonstrates that although AgentCAT relies on LLMs for semantic decision-making, its ultimate behavior successfully replicates the principle of Fisher information maximization in a statistical sense, achieving an organic unification of semantic reasoning robustness and mathematical optimality.

    
    
    

\subsection{Explanation of Knowledge Concepts(RQ3)}

Beyond the precision of ability estimation, the instructional logic and coherence of an adaptive assessment system during the interaction process are equally critical metrics for evaluating its usability. Traditional rule-based selection strategies tend to greedily search across the entire exercise pool for the exercise that yields the maximum statistical information. Although this approach may mathematically minimize estimation errors most rapidly, in real-world educational scenarios, it frequently causes the assessment trajectory to jump drastically between entirely unrelated knowledge concepts. Such a fragmented interactive experience not only violates the continuity principle of knowledge construction but also significantly increases the extraneous cognitive load on the examinee, ultimately leading to score losses caused by non-ability factors. To quantitatively evaluate the performance of selection strategies in maintaining knowledge consistency, we define a metric termed Knowledge Switch Rate (KSR), designed to measure the degree of concept jumping within the testing sequence. Formally, given a testing sequence $Q = \{q_1, q_2, ..., q_T\}$ of length $T$, where each question $q_t$ is associated with a primary knowledge concept $k_t$, the KSR is defined as:
\begin{align}
    \text{KSR} = \frac{1}{T-1} \sum_{t=2}^{T} \mathbb{I}(k_t \neq k_{t-1}),
\end{align}
where $\mathbb{I}(\cdot)$ is the indicator function. High KSR ($\approx 1.0$): Implies the test exhibits a Random Walk behavior, lacking logical coherence. Low KSR ($\to 0$): Implies the test is overly concentrated on a single concept, lacking exploration of comprehensive ability. Ideal CAT should maintain a moderate KSR. This signifies performing continuous diagnosis matched to the examinee's mastery level on a single knowledge concept, and then smoothly transitioning to logically related concepts only after information saturation is reached.

\begin{figure}[!tbp]
\centering

\includegraphics[width=0.5\linewidth]{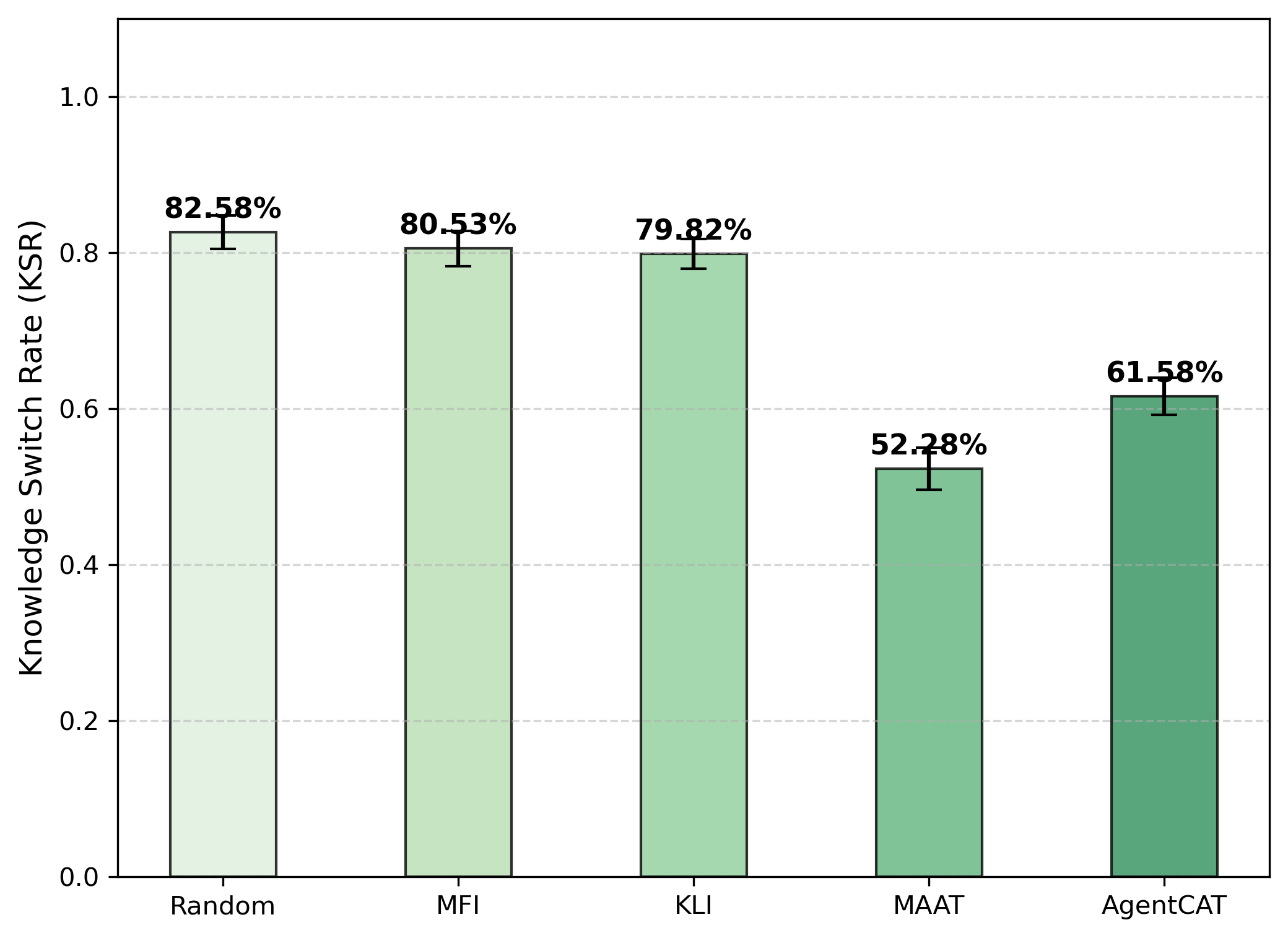}
\caption{Knowledge Coherence Comparison On XES3G5M (Lower is Better)}
\label{exp3_XES3G5M1}

\end{figure}

We conducted experiments on the XSE3G5M dataset. As illustrated in the Figure \ref{exp3_XES3G5M1}, the KSR of the baseline methods (Random/MFI/KLI) all hover around 80\%, exhibiting chaotic random jumps. In contrast, both MAAT (52.28\%) and AgentCAT (61.58\%) significantly reduced the switch rate. It is worth noting that MAAT's extremely low KSR is not entirely driven by pedagogical logic, but to a certain extent reflects a "local lock-in" of the algorithm. MAAT adopts a "Quality-then-Diversity" mechanism, causing its Diversity module to frequently select from highly homogenized candidate sets; this makes it difficult to break away from the current knowledge concept, ultimately leading the system into the trap of over-diagnosis. Conversely, AgentCAT's slightly higher KSR (61.58\%) indicates its superior global planning capabilities. It maintains local coherence while actively transitioning to new domains via the knowledge graph (Tier 3 strategy) once diagnostic information reaches saturation, thereby achieving a balance between "coherence" and "exploration" that better aligns with human pedagogical intuition.
 
\begin{figure}[!tbp]
\centering

\includegraphics[width=0.65\linewidth]{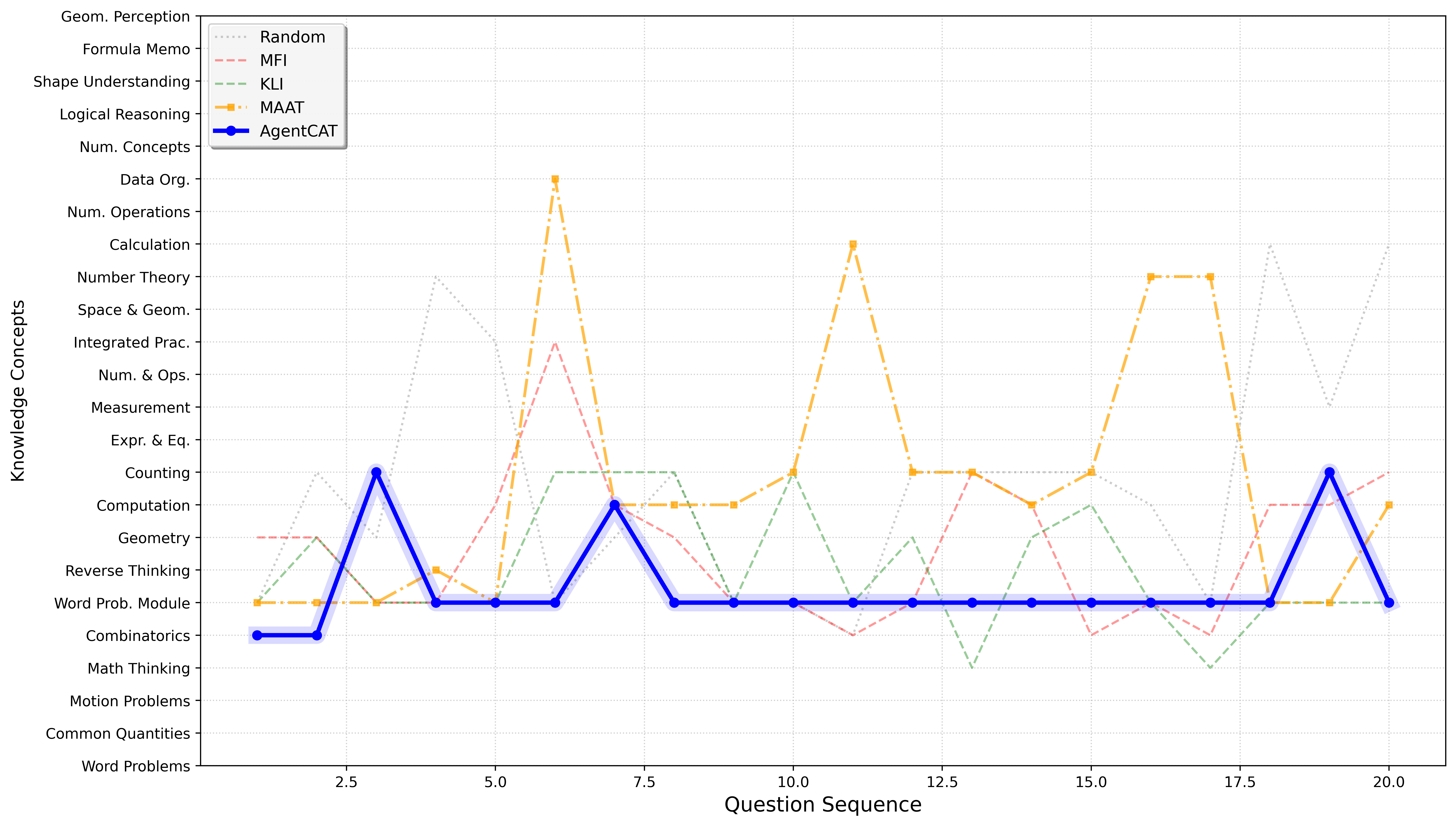}
\caption{The Testing Paths Of Each Strategy}
\label{exp3_XES3G5M2}
\end{figure}

The figure~\ref{exp3_XES3G5M2} further visualizes the instructional trajectories of each strategy: (1) The trajectories of Random, MFI, and KLI exhibit drastic up-and-down fluctuations, confirming their lack of intrinsic pedagogical guidance. (2) AgentCAT (blue solid line) demonstrates a unique staircase-like structure. Particularly in the interval from Q8 to Q18, the trajectory forms a stable straight line focused on the ``Word Prob. Module.'' This is not algorithmic stagnation, but a direct manifestation of AgentCAT executing an intra-bucket deep mining strategy—the system perceives that the examinee's cognitive state regarding this core module remains unclear, thus persisting in continuous longitudinal probing until sufficient diagnostic confidence is attained.

\subsection{Breaking Through The Limitations Of Static Evaluation(RQ4)}
In traditional CAT research, the assessment process typically relies strictly on the existing authentic response records within offline datasets. However, in authentic educational applications, this evaluation paradigm is constrained by the issue of partial labels in offline interaction logs, where a single examinee typically interacts with only a fraction of the exercises in the candidate pool. This implies that CAT can only conduct back-testing on the set of exercises that the examinee has previously "seen," whereas the vast majority of exercises remain unknown to the examinee. The role of these uninteracted exercises in proficiency assessment has long been overlooked. To this end, this experiment explores this unknown knowledge space, aiming to investigate whether the CAT system can still maintain robust pedagogical strategies and accurate ability estimation when an examinee faces exercises completely absent from their historical interaction data. Based on the XES3G5M dataset, we constructed a Golden Test Set specifically designed to break through data sparsity. Aiming to cover all knowledge concepts, this test set selects two exercises with the highest discrimination and moderate difficulty for each knowledge concept from the pool, thereby forming an exercise set with excellent diagnostic value. To validate the rationality and necessity of this testing environment, we conducted a quantitative comparison between the Golden Test Set and the examinees' historical response data across two dimensions: knowledge coverage and label sparsity.
As shown in Table \ref{tab:golden_set_necessity}: (1) Statistics reveal that the examinees' average unseen rate  for the exercises in the Golden Test Set reaches as high as 97.67\%. This indicates that the experiment successfully simulates the cold-start process in authentic scenarios, rather than a simple memory recall.
(2) The examinees' historical response data exhibits extreme sparsity, covering an average of only 21.04\% of the course knowledge graph. This demonstrates that traditional offline back-testing paradigms suffer from a massive diagnostic blind spot—the system's performance on nearly 80\% of the knowledge concepts is fundamentally unverifiable. In contrast, the Golden Test Set achieves a 99.91\% full-spectrum knowledge coverage.

\begin{table}[t]
    \centering
    \caption{Statistical Comparison of Data Sparsity and Knowledge Concept (KC) Coverage On XES3G5M.}
    \label{tab:golden_set_necessity}
    
    \small 
    
    \renewcommand{\arraystretch}{1.1}
    
    \begin{tabular}{cccc} 
        \toprule
        \textbf{Student ID} & \textbf{Unseen Q Rate} & \textbf{Hist. KC Cov.} & \textbf{Golden KC Cov.} \\
        \midrule
        7128  & 98.77\% & 22.36\% & 99.91\% \\
        9353  & 98.77\% & 20.62\% & 99.91\% \\
        4173  & 98.44\% & 20.81\% & 99.91\% \\
        ...   & ...     & ...     & ...     \\
        11829 & 96.72\% & 23.01\% & 99.91\% \\
        655   & 96.47\% & 23.65\% & 99.91\% \\
        8802  & 95.16\% & 25.39\% & 99.91\% \\
        \midrule
        \textbf{Average} & \textbf{97.67\%} & \textbf{21.04\%} & \textbf{99.91\%} \\
        \bottomrule
    \end{tabular}
\end{table}

\begin{figure}[htbp]
\centering

\includegraphics[width=0.5\linewidth]
{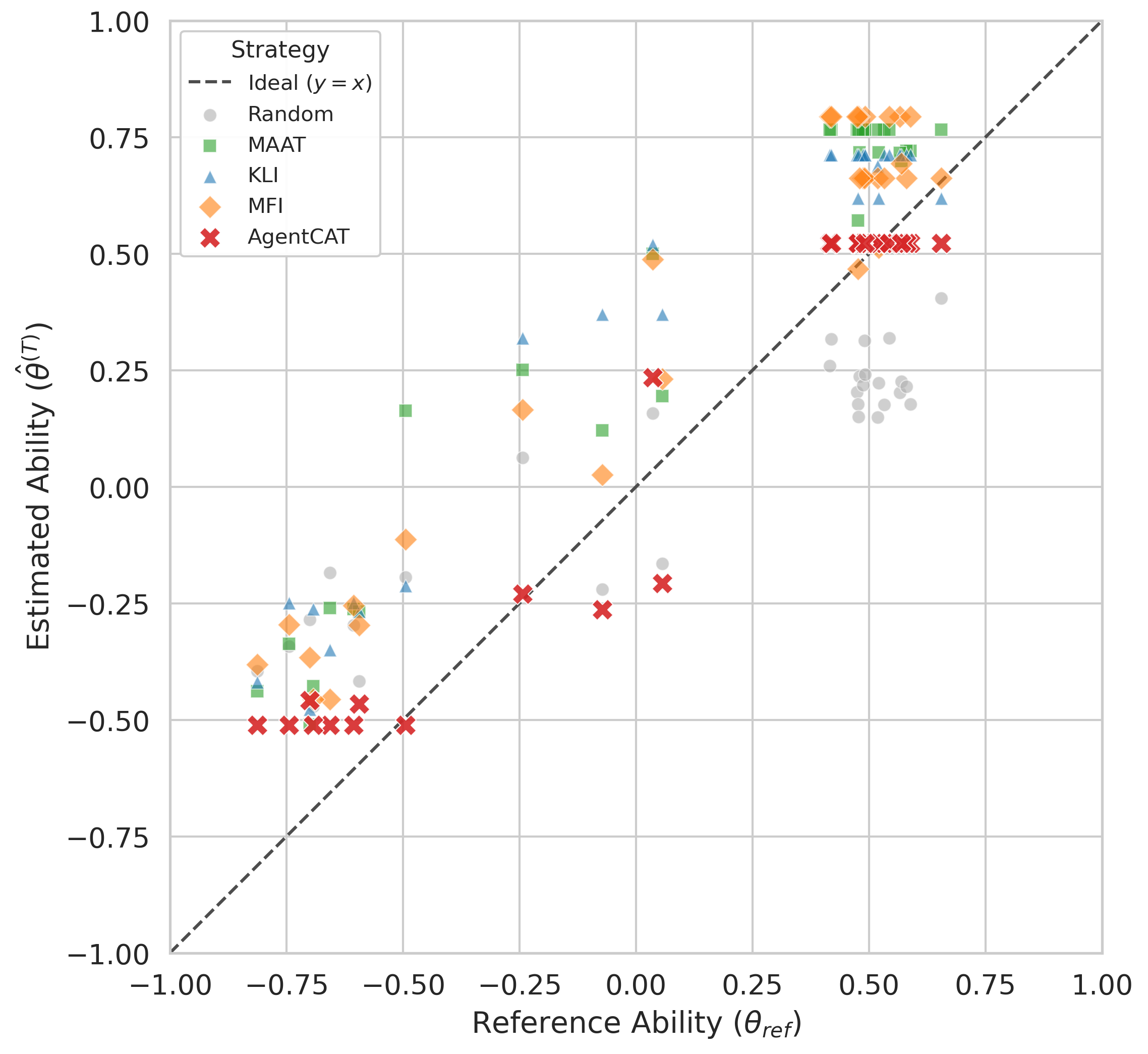}
\caption{Rank Consistency: Reference Ability vs Estimated Ability}
\label{exp4}
\vspace{-5mm}
\end{figure}

When confronted with 97.67\% of unobserved exercises, due to the absence of authentic response labels in the Golden Test Set, this experiment fully capitalizes on the advantages of AgentCAT as a high-quality simulation system. We utilize the examinee agent to generate responses on unknown exercises, and regard the ability estimation result after completing the full set of exercises as the examinee's reference ability ($\theta_{ref}$) within the full-spectrum space. Considering computational overhead and the constraints of authentic scenarios, we randomly sampled 50 exercises from the Golden Test Set to form the CAT candidate pool. Under identical settings, we compared the performance of AgentCAT against four traditional methods. Ultimately, the evaluation was conducted by calculating the MAE, RMSE, and the Spearman's rank correlation coefficient~\cite{liu2024computerized}—used to measure rank consistency—between the estimated ability $\hat{\theta}^{(T)}$ and the reference ability $\theta_{ref}$. As shown in Table \ref{tab:counterfactual_results}, in the counterfactual assessment scenario, traditional CAT methods suffer varying degrees of performance degradation in ability estimation precision. Conversely, AgentCAT significantly outperforms the baseline methods across all three evaluation metrics. Traditional statistical methods function as "parameter-sensitive" strategies that strictly rely on idealized IRT  probability assumptions. Consequently, they are highly susceptible to being misled by non-monotonic response behaviors generated by the examinee agent on unfamiliar exercises (e.g., random slips caused by semantic comprehension biases), thereby falling into local extrema of estimation bias.In contrast, AgentCAT does not merely rely on statistical parameters; instead, it executes "fault-tolerant decision-making" by leveraging semantic reasoning and the topological structure of the knowledge graph, effectively filtering out the interference of simulated noise. AgentCAT significantly outperforms the baseline methods across all three evaluation dimensions: it achieves the lowest errors in MAE and RMSE, while attaining a correlation coefficient of 0.8689—substantially higher than traditional methods. This conclusively demonstrates its exceptional generalization capability in complex, unknown environments.

\begin{table}[htbp]
    \centering
    \caption{Performance comparison on the Golden Test Set . }
    \label{tab:counterfactual_results}
    \renewcommand{\arraystretch}{1.1} 
    \setlength{\tabcolsep}{4mm} 
    
    \begin{tabular}{l c c c}
        \toprule
        \textbf{Method} & \textbf{MAE} $\downarrow$ & \textbf{RMSE} $\downarrow$ & \textbf{Correlation} $\uparrow$ \\
        \midrule
        Random   & 0.2864 & 0.3015 & 0.7535 \\
        MFI      & 0.2406 & 0.2738 & 0.7572 \\
        KLI      & 0.2573 & 0.2896 & 0.8046 \\
        MAAT     & 0.2749 & 0.3022 & 0.7454 \\
        \midrule
        \textbf{AgentCAT} & \textbf{0.0961} & \textbf{0.1281} & \textbf{0.8689} \\
        \bottomrule
    \end{tabular}

\end{table}
To more intuitively verify how well various methods discriminate the abilities of examinees, we plotted the scatter plot as illustrated in the figure \ref{exp4}. In the figure, the horizontal axis represents the examinee's reference true ability ($\theta_{ref}$), the vertical axis represents the $\hat{\theta}^{(T)}$ by each method, and the black dashed line ($y=x$) indicates the ideal zero-error state.  It can be observed that the red crosses (AgentCAT) adhere tightly around the diagonal. Whether in the low-ability zone ($\theta_{ref} < -0.5$) or the high-ability zone, AgentCAT's estimated values maintain high consistency with the reference values. In comparison, the scatter distributions of traditional methods are relatively loose, even exhibiting obvious outliers. This superior rank consistency implies that AgentCAT can not only accurately estimate an individual examinee's score but also precisely restore the examinee's relative ranking within the population. This holds immense practical value for selective examinations and stratified teaching.

\section{Analysis and Discussion}
Although AgentCAT demonstrates reliability in dynamic interaction and ability assessment, as an LLMs-based framework, its system overhead and result volatility are critical factors that must be considered in practical deployment and research.Within the multi-agent collaborative framework of AgentCAT, the primary computational overhead originates from the token consumption when the examinee agent generates long-text rationales and the selection agent conducts planning and retrieval. Taking the GLM-4-FlashX model as an example, according to the API call log statistics in our authentic testing environment, the average token overhead per API call within a continuously iterating long context is approximately 1200. For a complete adaptive testing sequence of length $T=20$, the total token consumption for the dual-end simulated interaction (including exercise retrieval and response simulation) of a single examinee is approximately 48,000. Under the current pricing standards of foundational large models, completing a full online dynamic assessment simulation for one examinee costs only about 0.005 RMB; even for the large-scale group simulation experiment with 30 concurrent users conducted in this study, the total inference cost is highly stably controlled within 0.15 RMB.

AgentCAT also exhibited certain failure cases during the experiments:~(1) Format Alignment Failure and Semantic Hallucination: In a minority of cases, the selection agent failed to correctly output the knowledge concept and bucket tags, or produced logical hallucinations. For instance, it might determine in its internal chain-of-thought that the examinee's ability is insufficient, yet select a high-difficulty feature bucket in its final output.~(2) Polarization and Local Oscillation in Ability Estimation: The experiments observed that in certain simulation trajectories, the system's ability estimation did not converge smoothly as expected, but instead exhibited severe overestimation or underestimation phenomena. This phenomenon likely stems from the distributional discrepancy between the simulated responses of large models and authentic human behaviors regarding the ``guessing'' and    ``slipping'' mechanisms. When the examinee agent, due to model hallucinations or over-reasoning, unexpectedly outputs the correct answer on a severely out-of-scope, highly difficult exercise, or provides an incorrect answer on a fundamental, error-prone exercise due to a logical lapse, the supervisor receives a massive penalty or reward gradient.

\section{Conclusion}
This paper proposes AgentCAT, a Large Language Model-based multi-agent collaborative simulation system designed to break through the constraints of traditional CAT research, which is chronically limited by static offline data and isolated optimization. Extensive experimental analysis robustly demonstrates the effectiveness and robustness of this simulation system in both ability estimation and selection strategies.
 
However, despite AgentCAT's superior overall performance, its current architecture still exhibits certain limitations, specifically in the design of the supervisor. Currently, when executing ability updates, the supervisor primarily relies on pre-defined mathematical rules. While this formulaic update paradigm is statistically interpretable, it may cause ability estimation to fall into local optima when faced with accumulated noise from long-term interactions or atypical response behaviors of the examinee. Future work will be dedicated to introducing Reinforcement Learning mechanisms to reconstruct the ability estimation module. We plan to design an IRT-based ability estimation agent to replace traditional formulaic numerical solvers. This agent will not rely on static mathematical formulas, but will instead model the ability update process as a sequential decision-making problem. It will be capable of intelligently perceiving subtle fluctuations in the examinee's cognitive state based on the current dialogue history and exercise features, dynamically adjusting the step size and direction of ability updates. Through this approach, we expect to achieve a paradigm shift in ability assessment from mechanical calculation to intelligent perception, thereby further enhancing the system's adaptability and assessment precision in complex, uncertain educational environments.


\bibliographystyle{ACM-Reference-Format}
\bibliography{samplebase}

\appendix

\

\end{document}